\documentclass[journal]{IEEEtran}
\usepackage{booktabs}
\usepackage{subfigure}
\usepackage{graphicx}
\usepackage{float}
\usepackage{verbatim}
\usepackage{graphicx,amssymb,mathrsfs,amsmath,latexsym,footnote}
\usepackage{epsfig}
\usepackage{epstopdf}
\usepackage{color,xcolor}
\usepackage{algpseudocode}
\usepackage[linesnumbered,ruled]{algorithm2e}
\usepackage{cite}
\usepackage[colorlinks,linkcolor=black,anchorcolor=black,citecolor=black]{hyperref}
\usepackage{multirow}
\usepackage{array}

\newcolumntype{C}[1]{>{\PreserveBackslash\centering}p{#1}}
\newcolumntype{R}[1]{>{\PreserveBackslash\raggedleft}p{#1}}
\newcolumntype{L}[1]{>{\PreserveBackslash\raggedright}p{#1}}

\makeatletter
\def\hlinew#1{%
  \noalign{\ifnum0=`}\fi\hrule \@height #1 \futurelet
   \reserved@a\@xhline}

\newcommand{\PreserveBackslash}[1]{\let\temp=\\#1\let\\=\temp}

\begin{document}
\title{UIU-Net: U-Net in U-Net for Infrared Small Object Detection}

\author{Xin Wu,~\IEEEmembership{Member,~IEEE,}
        Danfeng Hong,~\IEEEmembership{Senior Member,~IEEE,}
        Jocelyn Chanussot,~\IEEEmembership{Fellow,~IEEE}
\thanks{This work was supported by the National Natural Science Foundation of China under Grant 62101045 and Grant 42271350. This work was also supported by the MIAI@Grenoble Alpes (ANR-19-P3IA-0003) and the AXA Research Fund. (\emph{Corresponding author: Danfeng Hong.})}
\thanks{X. Wu is with the School of Computer Science (National Pilot Software Engineering School), Beijing University of Posts and Telecommunications, Beijing 100876, China. (e-mail: xin.wu@bupt.edu.cn)}
\thanks{D. Hong is with the Aerospace Information Research Institute, Chinese Academy of Sciences, Beijing 100094, China. (e-mail: hongdf@aircas.ac.cn)}
\thanks{J. Chanussot is with the Univ. Grenoble Alpes, CNRS, Grenoble INP, GIPSA-Lab, 38000 Grenoble, France, also with the Aerospace Information Research Institute, Chinese Academy of Sciences, Beijing 100094, China. (e-mail: jocelyn@hi.is)}
}


\markboth{IEEE Transactions on Image Processing,~Vol.~XX, No.~XX,~2022}%
{Shell \MakeLowercase{\textit{et al.}}: Bare Demo of IEEEtran.cls for IEEE Journals}

\maketitle

\begin{abstract}
\textcolor{blue}{This is the pre-acceptance version, to read the final version please go to IEEE Transactions on Image Processing on IEEE Xplore.}
Learning-based infrared small object detection methods currently rely heavily on the classification backbone network. This tends to result in tiny object loss and feature distinguishability limitations as the network depth increases. Furthermore, small objects in infrared images are frequently emerged bright and dark, posing severe demands for obtaining precise object contrast information. For this reason, we in this paper propose a simple and effective ``U-Net in U-Net'' framework, UIU-Net for short, and detect small objects in infrared images. As the name suggests, UIU-Net embeds a tiny U-Net into a larger U-Net backbone, enabling the multi-level and multi-scale representation learning of objects. Moreover, UIU-Net can be trained from scratch, and the learned features can enhance global and local contrast information effectively. More specifically, the UIU-Net model is divided into two modules: the resolution-maintenance deep supervision (RM-DS) module and the interactive-cross attention (IC-A) module. RM-DS integrates Residual U-blocks into a deep supervision network to generate deep multi-scale resolution-maintenance features while learning global context information. Further, IC-A encodes the local context information between the low-level details and high-level semantic features. Extensive experiments conducted on two infrared single-frame image datasets, i.e., SIRST and Synthetic datasets, show the effectiveness and superiority of the proposed UIU-Net in comparison with several state-of-the-art infrared small object detection methods. The proposed UIU-Net also produces powerful generalization performance for video sequence infrared small object datasets, e.g., ATR ground/air video sequence dataset. The codes of this work are available openly at \url{https://github.com/danfenghong/IEEE_TIP_UIU-Net}.

\end{abstract}

\begin{IEEEkeywords}
Infrared small object, deep learning, deep multi-scale feature, attention mechanism, local and global context information, feature interaction.
\end{IEEEkeywords}
%
\IEEEpeerreviewmaketitle

\section{Introduction}
\begin{figure}[!t]
    \centering
    \includegraphics[width=0.5\textwidth]{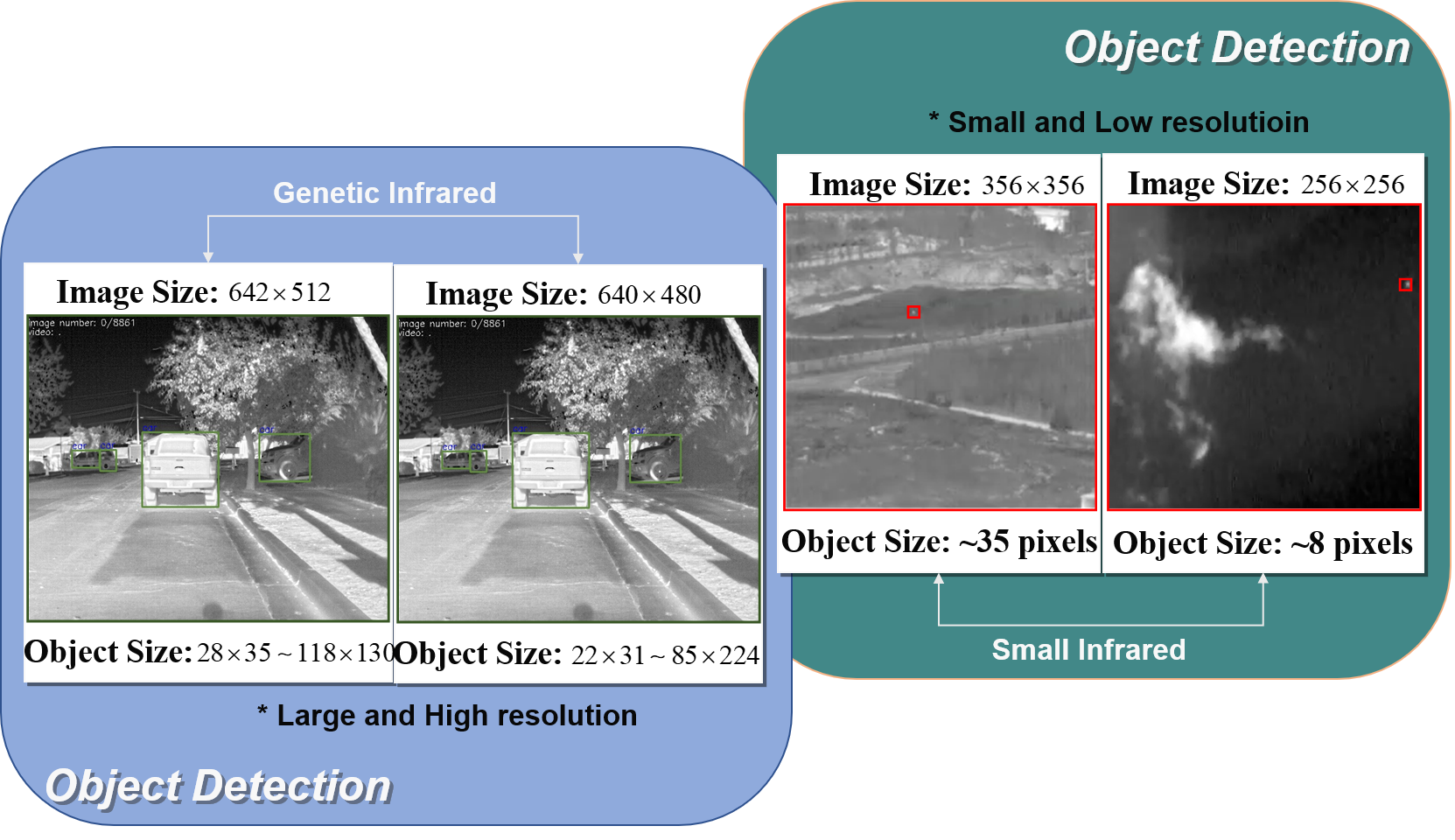}
    \caption{The visual difference between traditional infrared object detection and infrared small object detection.}
\label{fig: Vis_Diff}
\end{figure}

\IEEEPARstart{I}{nfrared} sensors \cite{Sta2016} are widely used in civil and military applications \cite{wu2022deep} since it is insensitive to the environment, illumination, occlusion, and other conditions. Infrared object detection mainly includes generic object detection (e.g., vehicle detection, pedestrian detection \cite{LSK2017} and re-identification \cite{CGAN2019}) and small object detection \cite{TNLRS2020}. The visual difference between them has been shown in Fig. \ref{fig: Vis_Diff}. Different from the generic objects, infrared images with small object acquisition often come from a far distance, and the size of objects usually covers less than $30\times 30$ pixels. This inevitably occurs in rescue and security missions, such as long-range maritime rescue missions of people or ships, black-flying drones, or airborne floating objects. Not only are these objects in small size, but they are often submerged in complex backgrounds, and lack color and texture information, making detection challenging. At present, there are two active research fields for infrared small object detection: single frame image object detection and video sequence image object detection. This paper focuses on the former.

\begin{figure}[!t]
    \centering
    \includegraphics[width=0.5\textwidth]{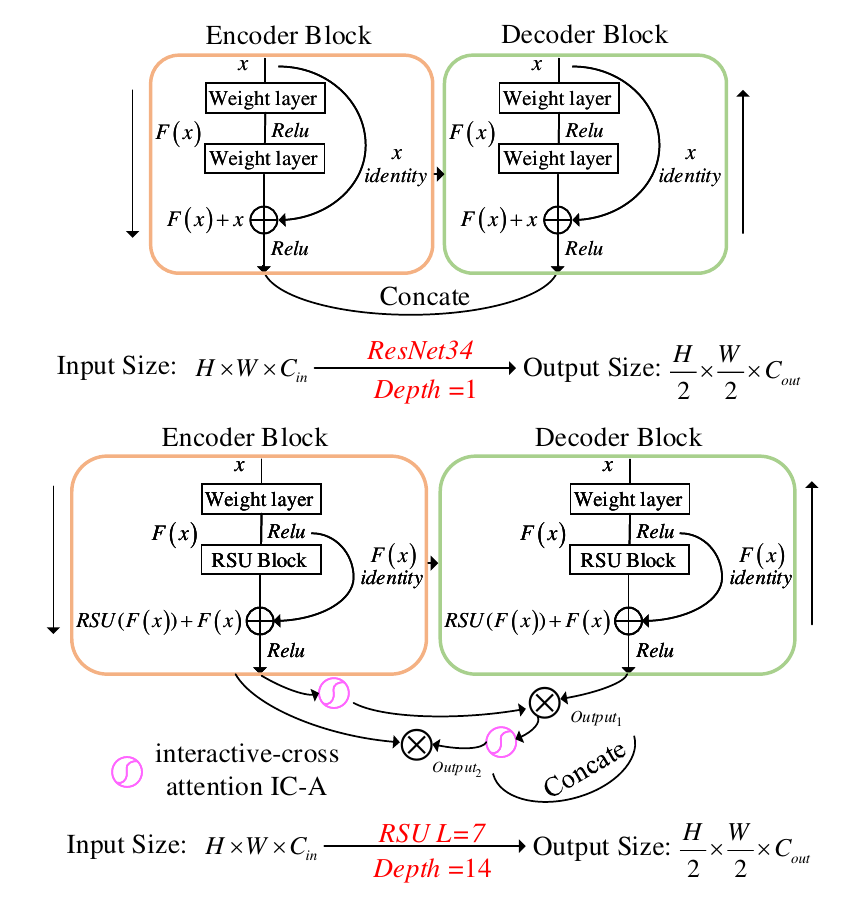}
    \caption{Depth comparison of classification backbone-based and RSU-based U-Net networks when the image resolution is reduced by $1/2$.}
\label{fig: ShortIn}
\end{figure}

Early common methods for single-frame infrared small object detection are model-based methods. It can be summarized as a filter-based method, human vision system-based method, and low-rank-based method. Among them, the filter-based method is just for single and uniform scenes \cite{bae2012edge}. The human vision system-based method is perfectly adequate for the object luminance is relatively large and presents a more obvious difference with the surrounding background, e.g., infrared patch image (IPI) \cite{gao2013infrared}, local contrast mechanism (LCM) \cite{chen2013local}, and multi-scale patch-based contrast measure (MPCM) \cite{wei2016multiscale}. The low-rank-based method, including local low-rank \cite{he2015small} and non-local low-rank \cite{TNLRS2020}, is clearly applicable to almost all kinds of complex and rapidly changing backgrounds, but in practice it requires acceleration by GPUs, etc., to make it meet the real-time needs. Recently, many new model-based methods have been developed, such as multi-scale gray difference weighted image entropy (MGDWIE) \cite{deng2016infrared} by weighting the local entropy, novel local contrast descriptor (NLCD) \cite{qin2019infrared} based on the facet kernel and random walker method, nonconvex method based on the partial sum of the tensor nuclear norm (PSTNN) \cite{zhang2019infrared}, generalized the IPI model called reweighted infrared patch-tensor (RIPT) \cite{dai2017reweighted} model, transformed domain filter-based methods nonnegativity-constrained variational mode decomposition (NVMD) \cite{wang2017infrared}, which achieves better detection performance. However, the model-driven method is susceptible to clutter and noise, reducing the robustness of the detection model. It also commonly fails to find an acceptable template or learn the objects' local contrast information with a complex background, or object modeling is heavily influenced by the model's hyperparameter, resulting in poor generalization performance.

In recent years, the explosion of data-driven machine learning-based methods \cite{hong2021graph,hong2021more}, especially deep learning methods, has rapidly made them the most widely used method for detecting small infrared objects. Wang et al. \cite{wang2017small} transferred a detection model trained by ImageNet Large Scale Visual Recognition Challenge (ILSVRC) data to detect infrared small objects. Nasser et al. \cite{nasrabadi2019deeptarget} proposed an automatic object recognition (ATR) framework by contacting two deep convolutional neural networks (DCNN) that were pre-trained on the ImageNet dataset. However, the detection performance for the infrared small objects will be severely limited if we direct migrate a pre-trained detection model based on natural scene images or direct train CNN with a typical downsampling mechanism. As a result, various networks have been proposed that are specifically for infrared small object detection. Fan et al. \cite{fan2018dim} designed a CNN-based architecture to learn suitable filters suitable for extracting the object and background sub-images in order to improve the detection performance of small and dim infrared object images. Lin et al. \cite{liangkui2018using} used a seven-layer network to detect infrared small objects by learning the synthesis data generated by oversampling. Wang et al. \cite{wang2019detection} developed a feature extraction backbone network called MNET for infrared small objects detection. By optimizing the target-to-clutter ratio  (TCR) the criterion to emphasize the representation of the infrared small object, McIntosh et al. proposed a TCR detection network (TCRNet) \cite{mcintosh2020infrared}. Hou et al. \cite{hou2021ristdnet} designed a robust infrared small object detection network (RISTDnet) by integrating handcrafted feature methods and convolutional neural networks to jointly learn the features of infrared small objects. By modeling the detection problem as an image-to-image translation problem, Zhao et al. \cite{zhao2020novel} developed an infrared small object detection method with a generative adversarial network (GAN). Heieh. et al. \cite{hsieh2021fast} proposed a high-speed detection method with a three-layer patch image model based on a layered gradient kernel. To address the problem of highly unbalanced foreground and background in infrared small object images, Zhao et al. \cite{zhao2019tbc} developed an infrared small object detection method (TBC-Net) with encoder-decoder by  adding high-level semantic constraint information of images.

Actually, modeling infrared small object detection as a semantic segmentation problem, rather than a typical object detection problem, helps better address the loss of model detection performance caused by the objects' small size. Unlike the general segmentation networks, e.g., semantic segmentation in natural scenes \cite{strudel2021segmenter}, organ segmentation for medical images \cite{zhang2021automatic}, the object shape needs to be finely and clearly segmented. However, due to their different imaging mechanism, infrared small targets present their morphology as the brightest/darkest regions with no obvious shape priors. Because the sensors record the thermal radiation energy emitted from the object at a large distance. As a result, the segmentation networks have to focus on small objects' saliency and discriminability characteristics. Dai et al. \cite{dai2021asymmetric, dai2021attentional} built a bottom-up attention mechanism module and embedded it into U-Net or FPN network structure. Similar to reference \cite{dai2021asymmetric, dai2021attentional}, Li et al. \cite{li2021dense} proposed the deny nested interactive module (DNIM) to integrate the context feature well using the U-Net segmentation network. However, almost all of the networks in the above work are based on various classification backbones with a typical downsampling scheme, such as Resnet-20 used in asymmetric contextual module (ACM) \cite{dai2021asymmetric} and attentional local contrast network (ALCNet) \cite{dai2021attentional}, and Resnet-18/34 used in DNIM \cite{li2021dense}. These classification backbones are generated by the Imagenet data, which are limited for the specific data distribution and spectrum. In addition, multiple downsampling in the network reduces feature resolution and loss of local contrast information, which is especially detrimental for infrared small objects. There is a worry that the continued use of these default backbone networks could not resolve the underlying problem of infrared small object detection.
\begin{figure*}[!t]
    \centering
    \includegraphics[width=1.0\textwidth]{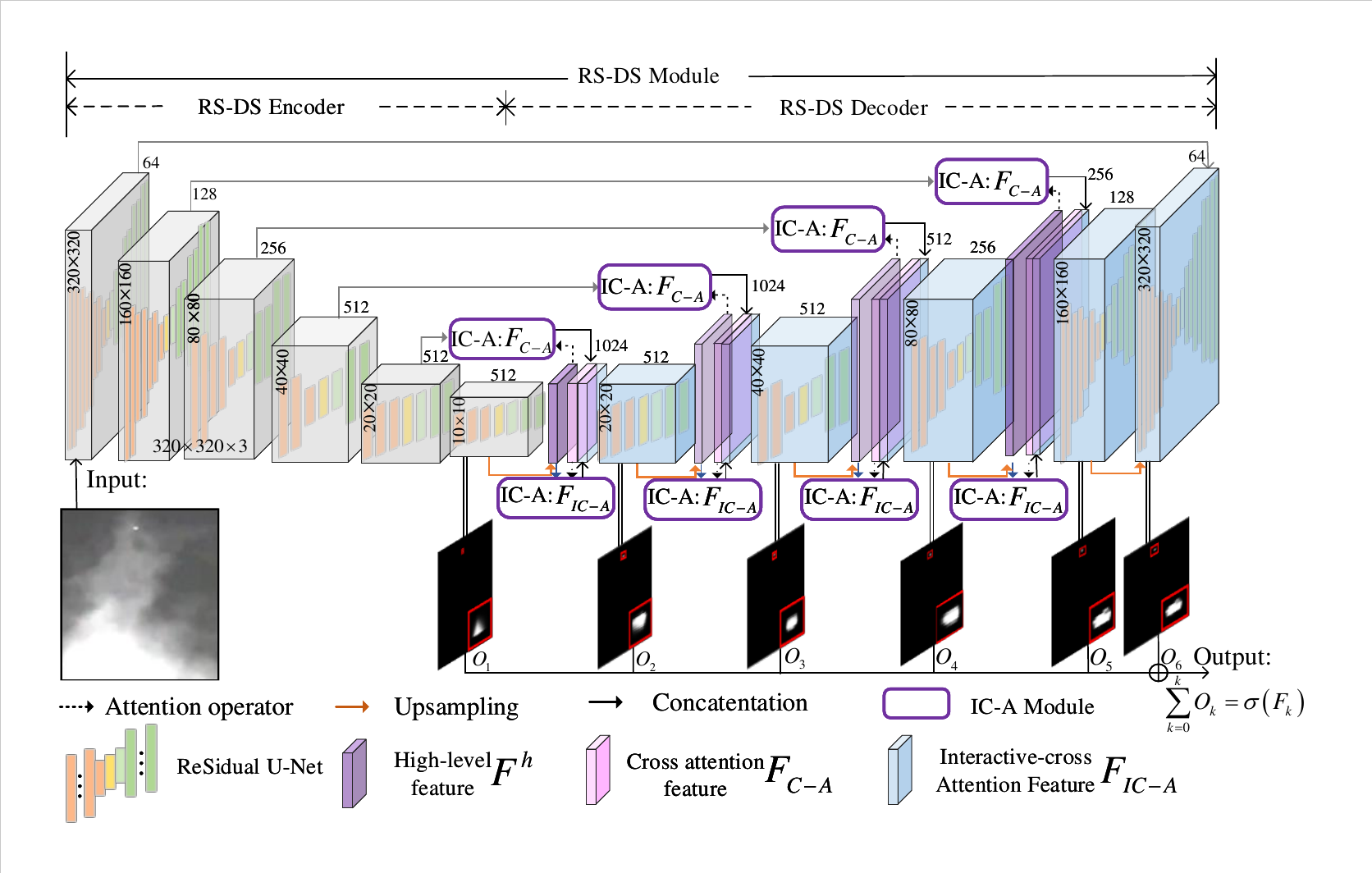}
    \caption{An overview of the proposed UIU-Net for infrared small object detection. It consists of two main modules: the resolution-maintenance deep supervision (RM-DS) module and the interactive-cross attention (IC-A) module. RM-DS improves global context representations by learning deep multi-scale features. IC-A encodes RM-DS features to further enhance local context representations. }
\label{fig:outline}
\end{figure*}

To this end, we proposed a simple and clear infrared small object detection framework by modeling infrared small object detection as a semantic segmentation problem, called U-Net in U-Net (UIU-Net). It is ideal for infrared object detection with small object sizes and limited data sizes. Specifically, it can be trained from scratch while improving the contextual representations at the global and local levels.  Firstly, the resolution-maintenance deep supervision (RM-DS) network is designed by integrating ReSidual U-blocks (RSU) into a deep supervision network to generate deep multi-scale and high-resolution features instead of pre-trained classification backbone networks. RM-DS resolves the contradiction between feature resolution and network depth. Moreover, it can also improve the objects' global context representation as the network grows deeper. Fig. \ref{fig: ShortIn} compares the depth of the classification backbone-based and RSU-based U-Net networks when the resolution of the image decreases by $1/2$. For further highlighting the contextual representations at the local levels, an interactive-cross attention (IC-A) module has been proposed and embedded in the RM-DS network. The IC-A module captures long-range dependencies between pixel-based objects by interactively cross-coding low-level details and high-level semantic features in place of the skip layer in U-Net. The actual output of UIU-Net is the fusion of the multi-layer output of the RM-DS network. 
In detail, our contributions in this paper can be summarized as follows.
\begin{itemize}
    \item We model infrared small object detection as a semantic segmentation problem and proposed an interactive-cross attention nested U-Net network that was trained from scratch, called U-Net in U-Net (UIU-Net for short).
    \item Resolution-maintenance deep supervision (RM-DS) network is proposed to learn deep multi-scale features to improve global context representations. 
 Interactive-cross attention (IC-A) module is designed to encode semantic features across low-level and high-level representations to further enhance the locally contextual contrast ratio.
    \item We evaluate the detection performance of the proposed UIU-Net on two different infrared single-frame image datasets, i.e., the SRIST data and the synthetic data, yielding significant advantages compared to several state-of-the-art detection methods. Furthermore, we also test the generalization performance on ATR ground/air video sequence dataset, and UIU-Net also shows its competition and superiority.
\end{itemize}

The rest of this paper is organized as follows. Section II briefly reviews object segmentation and attention mechanism methods. Section III detailly introduces the network architecture of the proposed UIU-Net, including the resolution-maintenance deep supervision module and interactive-cross attention module. Experiential results on two single frame image data and one video sequence data are analyzed in section IV. Section V gives a possible outlook for the future.

\section{Related Work}
\subsection{Object Segmentation}
Image segmentation is a pixel-level classification problem that involves dense image prediction. The explosion of convolutional neural networks (CNN) makes the image segmentation mission transform from patch classification \cite{hong2021multimodal}, fully convolutional networks (FCN) \cite{long2015fully}, encoder-decoder framework\cite{badrinarayanan2017segnet} to dilated/atrous framework \cite{wu2019fastfcn}. Subsequently, new semantic segmentation-based infrared small object detection networks, including ACM\cite{dai2021asymmetric}, ALCNet\cite{dai2021attentional}, DNIM \cite{li2021dense}, have begun arriving. This network's detection performance has improved tremendously compared to model-based and DL-based detection methods. However, these methods mostly rely on classification backbone networks with a typical downsampling scheme. Even novel down-sampling schemes updated by some researchers \cite{li2019scale,singh2018analysis}, the local and global contrast information of the infrared small object has still been ignored. 

Different from the above works, we design a multi-scale depth supervision structure to resolve the conflict between feature resolution and network depth, resulting in improved global and local context representation, as well as improved final detection performance.

\subsection{Attention Mechanism}
The attention mechanism was first proposed in the field of computer vision. Deepmind, a division of Google, created an attention mechanism for image classification in $2014$, allowing neural networks to pay more attention to relevant input regions while generating prediction missions. Subsequently, several types of research on attention mechanisms\cite{vaswani2017attention,guo2021beyond,hu2018squeeze,woo2018cbam,park2018bam} have also been proposed and widely used in many fields, including image classification \cite{hu2018squeeze,guo2021beyond}, semantic segmentation \cite{guo2021beyond}, object detection \cite{woo2018cbam}, and more, with good results. These attention mechanisms embedded in general networks (e.g., classification backbone) are not suited for detecting infrared small objects 
because small-sized objects are frequently missing in deep semantic feature maps.

Different from previous works, the proposed interactive-cross attention can first encode pixel-based local contextual features that enhance the detail information of infrared small objects, and the encoded features are derived from a resolution-maintaining deep supervision network that can learn multi-scale and global features. The discriminability of infrared small objects and backgrounds can be greatly improved by this integration.

\begin{figure*}[!t]
    \centering
    \includegraphics[width=0.9\textwidth]{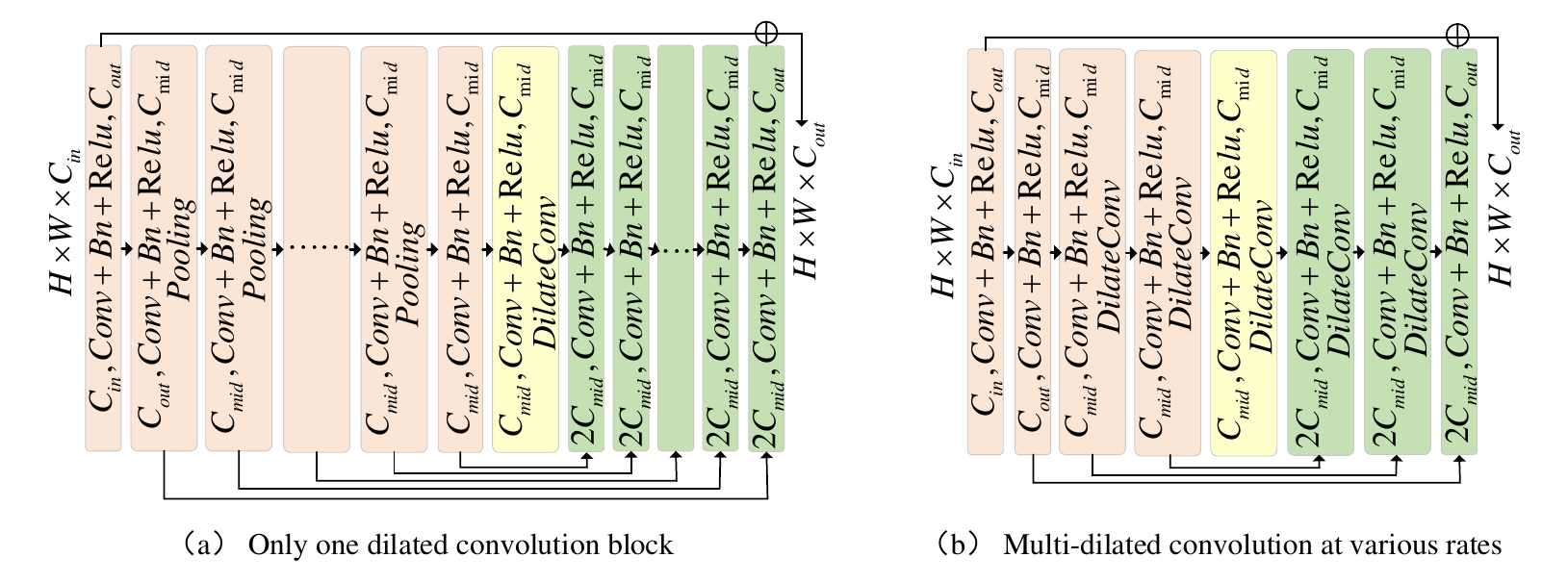}
    \caption{Illustration for the two modes of Residual U-block in resolution-maintenance deep supervision module.  (a) Only one dilated convolution in the last layer. (b) multi-dilated convolution at various rates in multiple layers.}
\label{fig:RSU}
\end{figure*}

\section{Methods}
In this section, we briefly review the simple but effective U-Net network for semantic segmentation and give a detailed discussion of the proposed UIU-Net. Fig. \ref{fig:outline} illustrates the overall network architecture of UIU-Net. UIU-Net begins with a resolution-maintenance deep supervision (RM-DS) network to learn deep multi-scale features while improving global context representation and then feeds them to the interactive-cross attention (IC-A) module to further encode object local context information. The IC-A module can also implement a seamless connection with the MD-DS network.

\subsection{Network Overview}
\textit{Review of U-Net Framework.} To begin, we briefly review the classical U-Net, which was initially developed for semantic segmentation, especially medical image segmentation. U-Net performs multi-scale prediction and provides finer segmentation features. The low-resolution feature map after multiple downsampling in the encoder stage could obtain the global semantic information of the segmented object. Skip connection in the decoder stage ensures the final reconstructed feature map merges more low-level features and also blends multi-scale features. However, multiple downsampling deteriorates feature resolution and object contrast, which yielded a very negative effect on infrared small object detection.

\textit{Network Overview.} To address the above problem, and ensure the detection performance of the infrared small objects. In this paper, we have proposed an interactive-cross attention-nested U-Net network with training from scratch, called U-Net in U-Net (UIU-Net). UIU-Net is not depending on a classic classification backbone network and is ideal for infrared small object detection. It begins with a resolution-maintenance deep supervision module that learns deep multi-scale features while improving global feature representation. The resolution maintenance here refers specifically to the feature learned throughout each stage of the encoder-decoder network. After that, the learned features from each stage are then fed into the interactive-cross attention module, which encodes objects' local features to increase the distinguishability ability of infrared small objects. Finally, the multiple intermediate supervision and the last layer are weighted and merged by minimizing a typical cross-entropy loss.

\subsection{Resolution-maintenance Deep Supervision (RM-DS) Module.}
Infrared images typically lack higher contrast due to their own imaging condition, resulting in a very low distinguishability between the object and background. This, coupled with the small size of the objects in the image, makes it difficult to precisely locate infrared small objects with a typical DL-based detection network or backbone-based segmentation network. In this section, we introduce a RM-DS network to overcome the above dilemma. RM-DS network is built on the base of U-Net. Multiple intermediate layers are utilized instead of just the last layer to obtain complete and distinguishable features, yielding better detection results.  The output $\mathcal{O}$ of RM-DS network is defined as 

\begin{equation}
\mathcal{O} = \sum\limits_{k = 0}^K \sigma( {F_k }),
\label{eq1}
\end{equation}
where $F_k$ denotes the $k$-th layer feature, and $\mathcal{O}$ is the final output of RM-DS network. $\sigma( \cdot )$ denotes $sigmoid$ activation function. 

In order to trade off the network depth and the feature resolution, the ReSidual U-blocks (RSU) module \cite{qin2020u2} is introduced as the backbone network. The detailed structure is illustrated in Fig. \ref{fig:RSU}. RSU module is a type of U-Net. The difference is that: 1) the RSU module takes intermediate feature maps rather than input images as input to learn and encode deep multi-scale features; 2) Dilated convolution is used to improve deep feature resolution in each layer. For shallow layers, only the last convolution is replaced with dilated convolution (see Fig. \ref{fig:RSU}(a)); However, for deep layers, all convolutions are replaced by dilated convolutions with varying dilated rates  (see Fig. \ref{fig:RSU}(b)), yielding low memory consumption due to the small size of deep feature maps; 3) Pooling operation has been introduced to the RSU module to reduce computing costs but has not been added to the additional RSU module to reduce feature loss. The feature of RSU backbone $U$ is defined as $ F_{U}= U(f(x))+f(x)$, where $f(x)$ is intermediate feature maps.
\begin{equation}
\mathcal{O} = \sum\limits_{k = 0}^K \sigma( {F_{U_k} }) = \sum\limits_{k = 0}^K \sigma( {U_k(f(x))+f(x)}),
\end{equation}
where $U_k(\cdot) (k = 1,2,\cdots, K)$ represents $K$ dilated-based convolutions modes, with just the last convolution replaced by dilated convolution and all convolutions replaced by dilated convolutions with different dilated rates. The particular use of which one is dependent on the situation.

Thanks to this design, the increase in network depth inside each stage has no effect on the feature resolution, and the growth in depth of each stage benefits from the growth in overall network depth. Meanwhile, an increase in network depth has enhanced the global context representation. \textbf{However}, only by MS-DS module not being able to locate infrared small object detection more precisely, especially under complex and ever-changing backgrounds.
\begin{figure}[!t]
    \centering
    \includegraphics[width=0.5\textwidth]{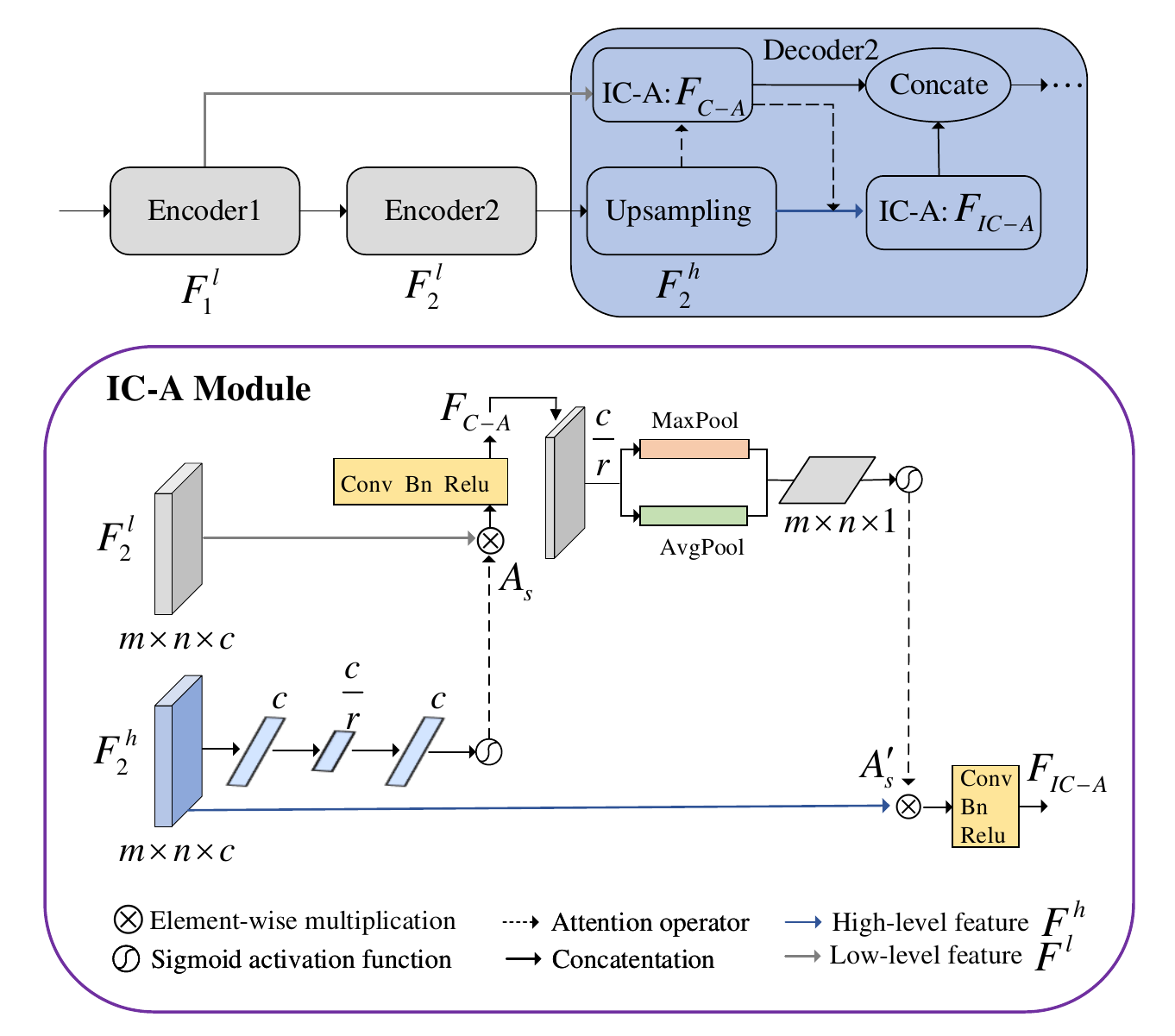}
    \caption{The detailed structure of interactive-cross attention module.}
\label{fig: IC-A}
\end{figure}

\subsection{Interactive-cross Attention (IC-A) Module}
In this section, an interactive-cross attention module is designed to encode local context representation and significantly improve detection performance. Figure \ref{fig: IC-A} depicts the detailed architecture. This module instead of the original skip connection in U-Net retains more context information from the decoder layer. Specifically, low-level detail and high-level semantic features are the coding objects of the interactive-cross attention module. Cross-channel attention and interactive-cross spatial attention are two suboutputs of the interactive-cross attention module. 

Each channel of high-level semantic features can be employed as specific object responses, and they are all associated with different semantic responses. We can focus on discovering the interesting infrared small objects, by making use of their interdependencies.

We define $U = [ {{u_1},{u_2}, \cdots ,{u_C}} ]$ as the encoder high-level features after ReSidual U-blocks \cite{qin2020u2} of deep network layer, where $u_c = F_{c}^{h} \in {\mathcal{R}^{W \times H}}$ is the $c$-th channel feature, and $C$ is the total number of channels. For these features, we first adopt \textit{AdaptiveAvgPool2d} (short for $\mathbf{F}_{ad}$) to traverse features 
\begin{equation}
\label{eq1}
\begin{aligned}
       \mathbf{z}_{c}= \mathbf{F}_{ad}(\mathbf{u}_c) = \frac{1}{{H \times W}}\sum\limits_{i = 1}^H {\sum\limits_{j = 1}^W {{u_c}\left( {i,j} \right)}},
\end{aligned}
\end{equation}
where $\mathbf{z}_c, c=1,2,\cdots, C$ stands the $c$-th channel. $H,W$ is the size of the high-level features. ${u_c}\left( {i,j} \right)$ is the feature of each location in the $c$-th channel. 

After that, the \textit{Excitation} operation (short for $\mathbf{F}_{ex}$) ($C \to {C \mathord{/ {\vphantom {C r}}
 \kern-\nulldelimiterspace} r} $, ${C \mathord{/ {\vphantom {C r}} \kern-\nulldelimiterspace} r} \to C$, r=4) is used to reshape high-level features ($F^h$) to minimize network parameters. 
 \begin{equation}
\label{eq1}
\begin{aligned}
       \mathcal{A}_\mathbf{s}= \mathbf{F}_{ex}(\mathbf{z}) = \sigma(\mathcal{B}(\mathbf{W}_2 \delta(\mathcal{B}(\mathbf{W}_1 \mathbf{z})))),
\end{aligned}
\end{equation}
where $\delta( \cdot ), \mathcal{B}(\cdot )$ are respectively the rectified linear unit (ReLU), and batch normalization (BN). ${\mathbf{W}_1} \in {R^{\frac{C}{r} \times C}}$ and ${\mathbf{W}_2} \in {R^{ C\times \frac{C}{r}}}$ stands for $C \to {C \mathord{/ {\vphantom {C r}}
 \kern-\nulldelimiterspace} r} $ and ${C \mathord{/ {\vphantom {C r}} \kern-\nulldelimiterspace} r} \to C$ \textit{Excitation} operation, respectively.
 
Finally cross channel features are a weighted sum of the features with low-level features, 
\begin{equation}
\label{eq1}
\begin{aligned}
       F_{C-A} = \mathcal{A}_\mathbf{s} \otimes F^l = \mathcal{A}_\mathbf{s} \otimes [U^l(f(x))+f(x)],
\end{aligned}
\end{equation}
where $\otimes$ denotes the element-wise multiplication, and $F^l$ denotes low-level features in the encoder. 

Considering infrared small object detection couldn't neglect the object's local detailed information to enhance the distinguishability between objects and background, interactive-cross spatial attention has been further designed. It first perform \textit{Excitation} operation ($\mathbf{F}_{ex}$) ($C \to {C \mathord{/ {\vphantom {C r}}
 \kern-\nulldelimiterspace} r} $, r=4) for $F_{C-A}$, and then aggregates it using both average-pooling ($\mathcal{P}_{avg}$) and max-pooling ($\mathcal{P}_{max}$) operations.
 \begin{equation}
\begin{aligned}
       \mathcal{A}_\mathbf{s}^\prime=
       &\delta({C^{3 \times 3}}( {{P_{avg}}( {{F_{C-A}^{1 \times 1}}} );{P_{\max }}({F_{C-A}^{1 \times 1}}}))),
\end{aligned}
\end{equation}
where  $C^{3 \times 3}$ and $C^{1 \times 1}$ are convolution operation, respectively. $F_{C-A}^{1 \times 1}$ stands for $F_{C-A}$ after $1 \times 1$ convolution.

The interactive-cross spatial attention features are a weighted sum of the features with high-level features, 
\begin{equation}
\begin{aligned}
       F_{IC-A} = \mathcal{A}_\mathbf{s}^\prime \otimes F^h = \mathcal{A}_\mathbf{s}^\prime \otimes [U^h(f(x))+f(x)],
\end{aligned}
\end{equation}
where $\otimes$ denotes the element-wise multiplication, and $F^h$ denotes high-level features in the encoder. 

Finally,  the output of UIU-Net is defined as 
\begin{equation}
\begin{aligned}
       \mathcal{O} = \sum\limits_{k = 0}^K \delta( {f(U_k) }) = \sum\limits_{k = 0}^K \delta( [F_k^{C-A}, F_k^{IC-A}])
\end{aligned}
\end{equation}

\section{Experiments}
\subsection{Dataset Description}
\subsubsection{\textbf{Single-frame InfraRed Small Target detection (SIRST) data}\cite{dai2021asymmetric}}
The SIRST dataset\footnote{\url{https://github.com/YimianDai/open-acm}} is a small open single-frame infrared small object detection dataset by selecting images from a sequence, which was designated as public by the University of Arizona, in 2020. It contains $427$ representative images of different scenarios from hundreds of real-world videos for different scenarios. These images were captured at short-wavelength, mid-wavelength, and $950$nm wavelengths. They are annotated with five different forms to support the model of detection task, and segmentation task. Most small infrared objects in this dataset are very dark, buried in a complicated background, and there is a lot of clutter. Furthermore, only $35$\% of the objects in this dataset belong to the brightest pixels. In the experiment, we randomly add $80$ representative images from synthetic multiple scenes infrared small target dataset (MSISTD) \footnote{\url{https://github.com/Crescent-Ao/MSISTD}} for model verification, and $20$ true images from SIRST for quantitative analysis.

\subsubsection{\textbf{Synthetic Infrared Small Target detection data}\cite{wang2019miss}}
The synthetic dataset\footnote{\url{https://github.com/wanghuanphd/MDvsFA_cGAN}} is a large open and synthetic single-frame infrared small object detection dataset. It is built up through the real infrared small object or the two-dimensional Gaussian function randomly overlaid on the high-resolution natural scene images from the Internet. Among them, the real infrared small object is selected from $11$ real sequences called ``AllSeqs''and $100$ individual infrared images called ``Single''. The synthetic dataset was designated as public by the Nanjing University of Science \&
Technology and the University of Sydney in $2020$. At present, only the configuration I mentioned in \cite{wang2019miss} has been opened and available. In this part dataset, the ``Single'' serves as the test set, while the ``AllSeqs'' and synthetic images serve as the training set. Most small infrared objects in this dataset are buried in a complicated background, and their labels are slightly larger than the actual object location.

\subsubsection{\textbf{ATR ground/air background infrared detection and tracking data }\cite{hui2020data}}
The ground/air data\footnote{\url{https://www.scidb.cn/en/detail?dataSetId=720626420933459968&version=V1&dataSetType=journal&tag=2&language=zh_CN}} is an infrared detection and tracking dataset with one or more fixed-wing UAV objects that differ in size, interference, storage and mobility, and other properties. It was designated as public by ATR Key Laboratory, College of Electronic Science and Technology, National University of Defense Technology. The data was collected between 2017 and 2019, and the images were obtained at a wavelength of 3-5um with a spatial resolution of 10-100m. Sky, ground, and a variety of other sceneries were used to collect this data, which totaled $22$ data segments, $30$ traces, $16,177$ frames, and $16,944$ objects. Each object corresponds to a labeled location, and each data segment corresponds to a labeled file. Infrared objects in this dataset are buried in a complex and shifting background, such as sky, ground, air-land junction, and so on.

\subsubsection{Implementation Details.} We conduct experiments on our proposed UIU-Net on the PyTorch platform with an NVIDIA GeForce GTX 1080 (8GB memory). During the network training,
each image is resized to $320\times 320$ by image scale and crop, and even color apace. For the input image size with $320\times 320$ pixels with $3$ bands, the computational cost of the UIU-Net in terms of Floating-point Operations Per Second (Flops) and parameters (Params) are 33.64G and 50.54M, respectively. The Adam optimizer is used for network optimization with the batch size $3$ and epoch number $500$.

\subsection{Evaluation Metrics}
In this paper, we model infrared small object detection as a semantic segmentation problem. The detection performance is evaluated by computing three commonly-used indices, i.e., Intersection over union (IoU), normalized intersection over union (nIoU), and the receiver operating characteristic (ROC) curve. In general, the higher values  for the three indices means better detection performance in the infrared small object detection task.

1) \textbf{\textit{Intersection over Union (IoU):}} IoU is widely used for semantic segmentation. It is calculated by the intersection of the real and predicted values of the pixel divided by their union. 
\begin{equation}
\begin{aligned}
IoU = \frac{{{A_{{\mathop{\rm int}} {\rm{er}}}}}}{{{A_{all}}}}= \frac{{\sum\limits_i^N {\sum\limits_m^M} {TP_m^i} }}{{\sum\limits_i^N {\sum\limits_m^M} T_m^i  + \sum\limits_i^N {\sum\limits_m^M} P_m^i  - \sum\limits_i^N {\sum\limits_m^M {TP_m^i} }}},
\end{aligned}
\label{IoU}
\end{equation}
where $M$ is the object number of each image/sample, and $N$ is the total sample number of the testing set. ${{A_{{\mathop{\rm int}} {\rm{er}}}}}$ and ${{A_{all}}}$ stands for the summation result for the intersection and union of the all real and predicted objects in the testing set. $T$, $P$, and $TP$ stand for reference label, predict the result, and true positive pixel numbers, respectively. the $i,m$ in $TP_m^i$, $T_m^i$, $P_m^i$, is the $m$-th object in $i$-th sample. 

\begin{table}[!t]
\renewcommand\arraystretch{1.5}
\centering
\small
\caption{The effect of different backbone networks embedded in UIU-Net. The best results are shown in bold.}
\resizebox{0.5\textwidth}{!}{
\begin{tabular}{c|c|c|c|c||c|c}
\toprule[1.5pt]
\multirow{2}{*}{No.} &\multirow{2}{*}{Baseline} & \multirow{2}{*}{backbone} &\multicolumn{2}{c||}{SIRST dataset} &\multicolumn{2}{c}{Synthetic dataset}\\\cline{4-7}
 && & IoU & nIoU  & IoU & nIoU\\\hline
1&U-Net&ResNet & 0.6178 & 0.6378 &0.4335  &0.4032\\
2&U-Net&RSU & \bf 0.7825 &  \bf 0.7515 & \bf 0.4773 &  \bf 0.4721\\ 
\bottomrule[1.5pt]
\end{tabular}
}
\label{tab:block}
\end{table}

\begin{table}[!t]
\renewcommand\arraystretch{1.5}
\centering
\small
\caption{The effect of different components in the UIU-Net with input in $320 \times 320$. The best results are shown in bold.}
\resizebox{0.5\textwidth}{!}{
\begin{tabular}{c|c|c|c|c|c||c|c}
\toprule[1.5pt]
\multirow{2}{*}{No.} &\multirow{2}{*}{Baseline} & \multirow{2}{*}{RSU} & \multirow{2}{*}{IC-A} &\multicolumn{2}{c||}{SIRST dataset} &\multicolumn{2}{c}{Synthetic dataset}\\\cline{5-8}
 && & & IoU & nIoU & IoU & nIoU \\\hline
1&U-Net& $\checkmark$& & 0.7330 &  0.7099  & 0.4100 &  0.3878\\
2&U-Net&\checkmark&\checkmark & \bf 0.7825 & \bf 0.7515 & \bf 0.4773 & \bf 0.4721\\ 
\bottomrule[1.5pt]
\end{tabular}
}
\label{tab:module}
\end{table}
\begin{figure*}[!t]
\centering\includegraphics[width=1.0\textwidth]{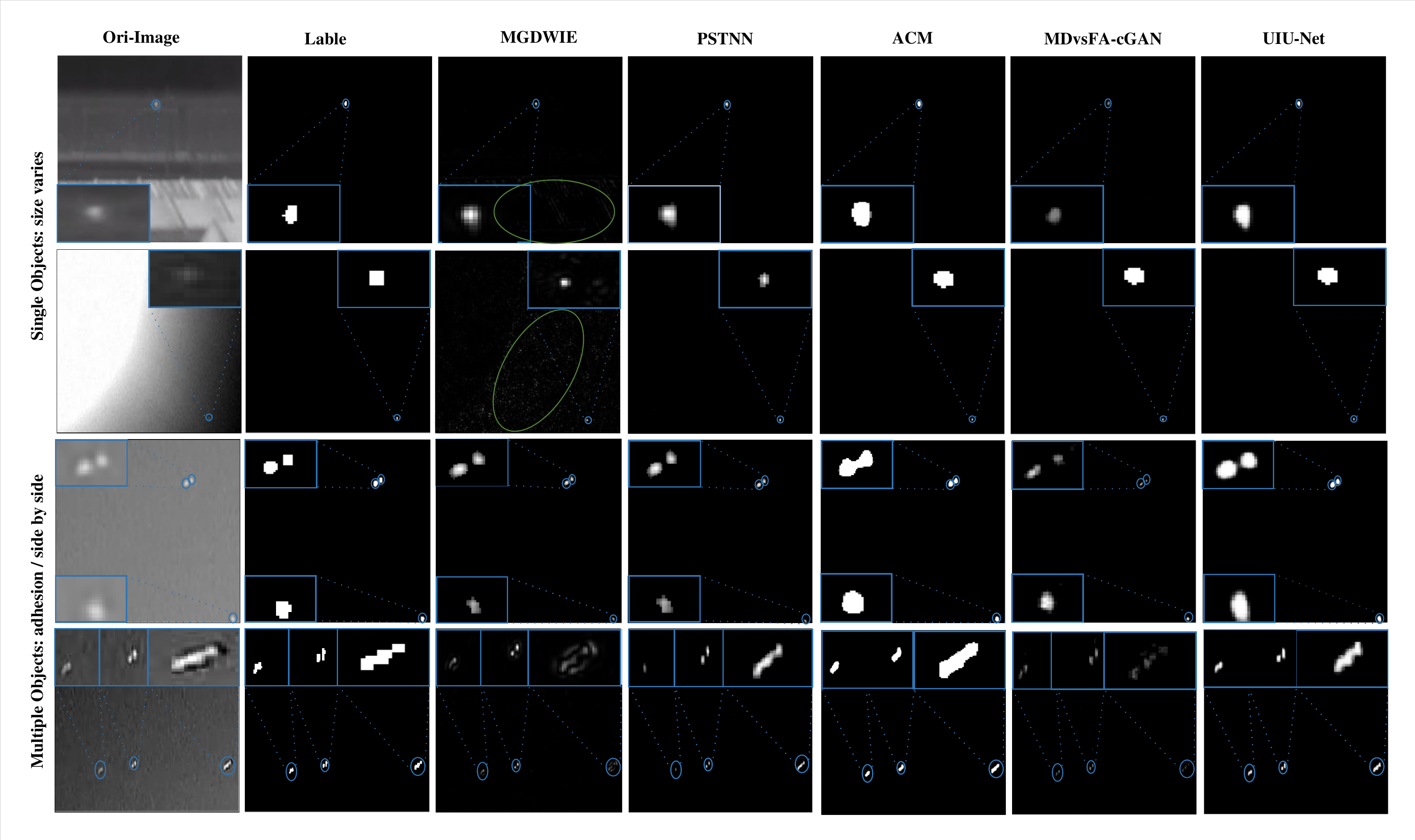}
\caption{Visual examples of some representative methods for SIRST dataset. The blue and green circles represent true positive and false positive objects, respectively. Inside the blue rectangle is the true positive objects after zooming in to be more clearly distinguish the detection accuracy of different methods.}
\label{fig:visSirst}
\end{figure*}

2) \textbf{\textit{normalized Intersection over Union (nIoU):}} signal-to-noise ratio (SCR) is calculated by the object and its surrounding pixels, which is widely used in the traditional method, is infinite in deep learning-based semantic segmentation. To better balance model-driven and data-driven methodologies, ref \cite{dai2021asymmetric} introduced the normalized Intersection over Union (nIoU) as a successor for the IoU. It has to be defined as

\begin{equation}
\begin{aligned}
       nIoU = \frac{1}{N}\sum\limits_i^N {\frac{{TP\left[ i \right]}}{{T\left[ i \right] + P\left[ i \right] - TP[i]}}},
\end{aligned}
\label{nIoU}
\end{equation}
where $N$ is still the total sample number of the testing sets. $\frac{{TP\left[ i \right]}}{{T\left[ i \right] + P\left[ i \right] - TP[i]}}$ stands for the IoU of each sample. $T$, $P$, $TP$ are same to Equ. \ref{IoU}, and $i$ in $TP[i]$, $T[i]$, $P[i]$, is the $i$-th sample.

3) \textbf{\textit{Receiver Operating Characteristic (ROC) curve:}} ROC presents the shifting trends between false positive rate (FPR) and true positive rate (TPR). Unlike IoU, which only reflects the segmentation effect under a fixed threshold, it indicates the total effect under a sliding threshold.

\begin{equation}
\begin{aligned}
       TPR_{thr} = \frac{{TP_{thr}}}{P_{thr}}, \;\; FPR_{thr} = \frac{{FP_{thr}}}{P_{thr}},
\end{aligned}
\end{equation}
where $TP$, $P$ are same to Equ. \ref{IoU} and Equ. \ref{nIoU}. $FP$ is false positive pixel numbers, and $thr$ stands for the threshold variable.


\subsection{Ablation Study} 
In this section, two ablation experiments for the two infrared small object datasets are listed to assess the effectiveness of the proposed UIU-Net. In detail, we investigated 1) the comparative detection performance of UIU-Net under different backbone networks, and 2) the individual contributions of each module in UIU-Net. For each part of the ablation study, we rigorously retrained the whole UIU-Net with the same parameter settings. 

\textit{1) the comparative detection performance of UIU-Net under different backbone networks.} We compare the IoU and nIoU values of U-Net with a classical residual backbone and multi-scale ReSidual U-block module in Table \ref{tab:block}. It shows that the IoU value increased from 0.6178 to 0.7825 on the SIRST dataset and from 0.4335 to 0.4773 on the synthetic dataset. For the nIoU value, the SIRST dataset improves by $\sim$ 0.12 to 0.7515, and the synthetic dataset improves by $\sim$ 0.07 to 0.4721. The above analysis clearly demonstrates the efficacy of the multi-scale residual backbone for infrared small object detection, and its ability to learn deep high-resolution features can greatly compensate for tiny object loss or poor feature representation.

\begin{table*}[!t]
\renewcommand\arraystretch{1.5}
\centering
\small
\caption{Quantitative comparison of different methods in terms of IoU and nIoU value on SIRST infrared small object datasets. The best results are shown in bold.}
\resizebox{1.0\textwidth}{!}{
\begin{tabular}{c|c|c|c|c|c|c|c|c|c|c}
\toprule[1.5pt]
\multirow{3}{*}{Methods}&\multicolumn{6}{c|}{Model-driven Method} & \multicolumn{4}{c}{Data-driven Method}\\ \cline{2-11}
&\multicolumn{10}{c}{Train 407  Test  20} \\ 
\cline{2-11}
 &NLCD \cite{qin2019infrared}  &PSTNN\cite{zhang2019infrared} &IPI\cite{gao2013infrared} &RIPT\cite{dai2017reweighted} &MGDWIE\cite{deng2016infrared} &NVMD\cite{wang2017infrared} & ACM\cite{dai2021asymmetric} & MDvsFA-cGAN\cite{wang2019miss} & TBC-Net\cite{zhao2019tbc} & UIU-Net \\
 \hline
IoU &0.2264&0.5121&0.4227&0.3080&0.1969&0.2034&  0.6178& 0.4114&0.6068&\bf 0.7825 \\
nIoU &0.3591&0.5729&0.4980&0.3587&0.3304&0.4571&  0.6378& 0.5653 & 0.6299 & \bf 0.7515\\
\bottomrule[1.5pt]
\end{tabular}
}
\label{tab:QuaSIRST}
\end{table*}
\begin{figure}[!t]
    \centering
    \includegraphics[width=0.5\textwidth]{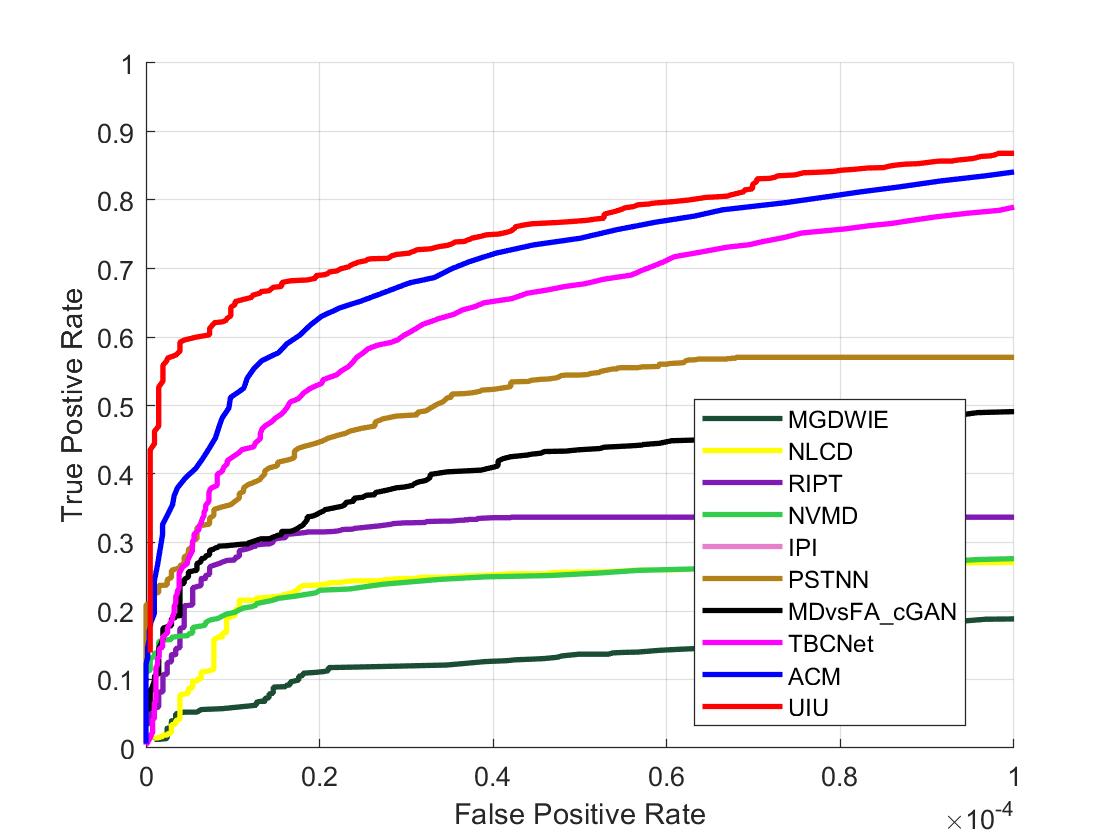}
    \caption{ROC Curve for the SIRST dataset. It's worth noting that the IPI ROC result is poor enough to be displayed.}
\label{fig:SIRROC}
\end{figure}

\textit{2) the individual contributions of each module in UIU-Net.} We compare the IoU and nIoU values of two incremental in Table \ref{tab:module}. For the incremental module (U-Net + RSU), the IoU value is increased from 0.7330 to 0.7825 on the SIRST dataset and from 0.4100 to 0.4773 on the synthetic dataset. For the nIoU value, the SIRST dataset improves by $\sim$ 0.05 to 0.7515, and the synthetic dataset improves by $\sim$ 0.09 to 0.4721. The above analysis clearly demonstrates the efficacy of the incremental module for the proposed UIU-Net. It also indicates that integrating interactive-cross encoder to deep high-resolution and multi-scale features is very necessary. 

\subsection{Results and Analysis on the SIRST Data}

Table \ref{tab:QuaSIRST} lists the detection accuracy of nine infrared detection methods, including infrared patch image (IPI) \cite{gao2013infrared}, local contrast measure (LCM) \cite{chen2013local}, reweighted infrared patch-tensor (RIPT) \cite{dai2017reweighted}, multiscale gray difference
weighted image entropy (MGDWIE) \cite{deng2016infrared}, nonnegativity-constrained variational mode decomposition (NVMD) \cite{wang2017infrared}, partial sum of tensor nuclear norm (PSTNN) \cite{zhang2019infrared}, novel local contrast descriptor (NLCD) \cite{qin2019infrared}, asymmetric contextual modulation (ACM) \cite{dai2021asymmetric}, Miss Detection vs. False Alarm-Conditional generative adversarial network (MDvsFA-cGAN) \cite{wang2019miss}, and lightweight convolutional neural
network (TBC-Net) \cite{zhao2019tbc}. in terms for two commonly-used indices IoU and nIoU, while Fig. \ref{fig:visSirst} shows four example methods of the corresponding detection results and the reference label.
\begin{figure*}[!t]
\centering\includegraphics[width=1.0\textwidth]{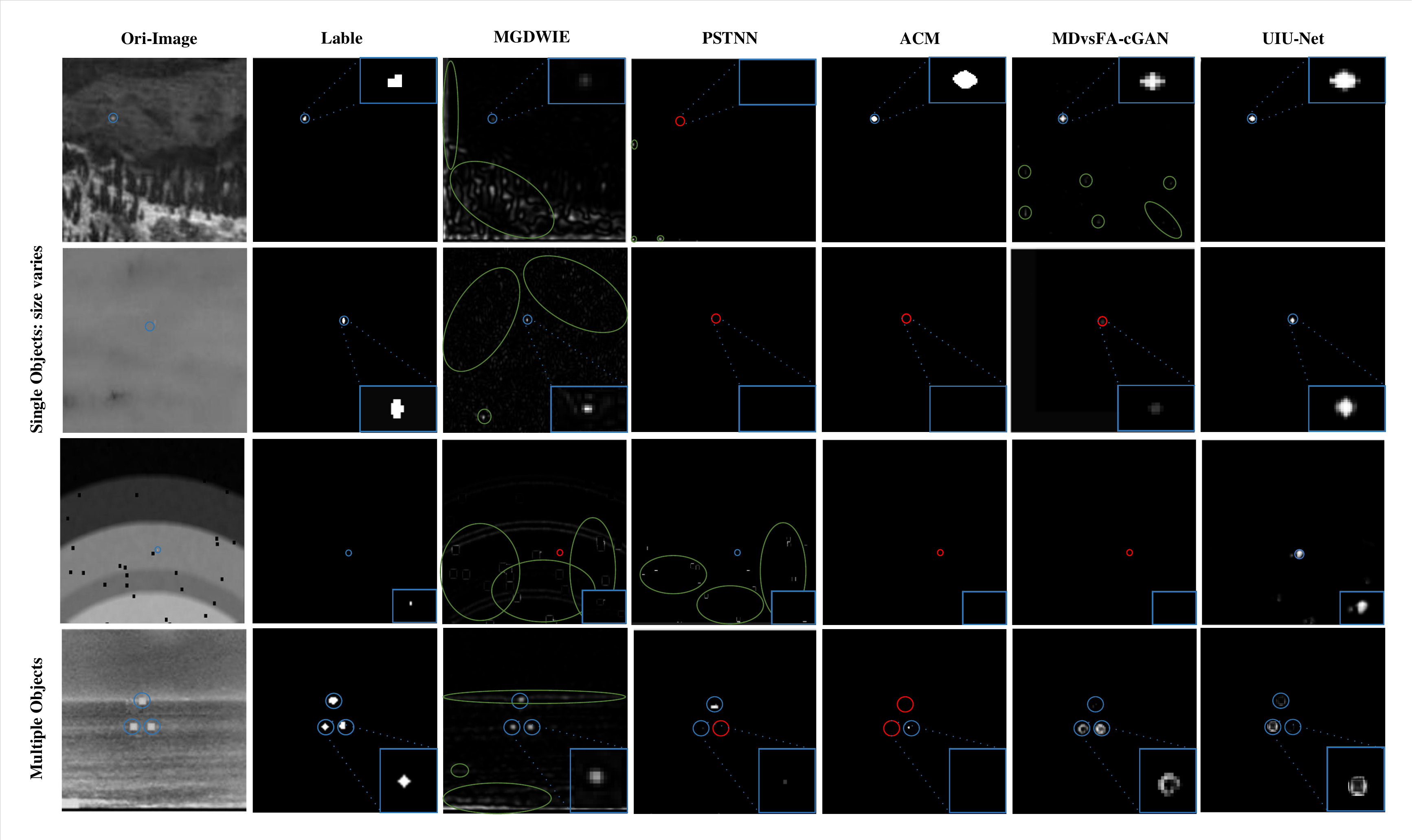}
\caption{Visual examples of some representative methods for the synthetic dataset. The blue, green, and red circles represent true positive, false positive objects, and missed detection objects, respectively. Inside the blue rectangle is the specific objects after zooming in to more clearly distinguish the detection accuracy of different methods.}
\label{fig:visSys}
\end{figure*}

\textit{1) Quantitative Comparison.} Overall, benefiting from the feature representation ability, the detection results using a data-driven network are dramatically higher than those model-driven methods, and the proposed data-driven network outperforms other compared methods, which obtains the best IoU and nIoU values in infrared small object detection. For the SIRST dataset with small and real data quantities, the UIU-Net learned deep multi-scale and high-resolution features and focused on the global and local contrast. This, to some extent, yields better detection performance and lower miss detection rates. ACM and TBC-Net have relied on the classification backbone trained by the ImageNet dataset, which makes the network mainly focus on the objects in SIRST with the same distribution as ImageNet rather than on the SIRST data itself, thus limited to the detection performance and also failing to fundamentally improve the detection performance. MDvsFA-cGAN model has balanced the missed detection rate and false alarm rate by adversarial generation network. However, the higher complexity of the MDvsFA-cGAN model cannot be applied to the small data of SIRST, making IoU and nIoU values lower than other data-driven methods.

Different from the IoU and nIoU values with fixed thresholds, we also present the ROC curves for the above methods with dynamic thresholds in Fig. \ref{fig:SIRROC}. As can be observed, the proposed method still performs the best. More specifically, IPI has displayed in a specific coordinate interval due to the low quality of the data.

\textit{2) Visual Comparison.} Some visualization examples of the SIRST dataset have shown in Fig. \ref{fig:visSirst}, including single objects and multiple objects. Obviously, the model-driven methods with limited feature expressiveness yielded a high rate of missing detection. The visual effect is that detected objects have fewer pixel values than the reference label. The other three data-driven methods with feature auto-learning obtained better detection results, which is consistent with the quantitative results in Table \ref{tab:QuaSIRST}. Specifically, the proposed method yielded optimal visual results, particularly for tightly connected multiple objects (see the blue circles in the third and fourth row of Fig. \ref{fig:visSirst}). But for ACM and MDvsFA-cGAN, their visual results show adhesion or pixel loss (as shown in the second and third-column in Fig. \ref{fig:visSirst}). Especially for MDvsFA-cGAN, which is trained by a small dataset (SIRST), the visual results are even worse than the model-based PSTNN method.

\begin{table*}[!t]
\renewcommand\arraystretch{1.5}
\centering
\small
\caption{Quantitative comparison of different methods in terms of IoU and nIoU value on the Synthetic datasets. The best results are shown in bold.}
\resizebox{1.0\textwidth}{!}{
\begin{tabular}{c|c|c|c|c|c|c|c|c|c|c}
\toprule[1.5pt]
\multirow{3}{*}{Methods}&\multicolumn{6}{c|}{Model-driven Method} & \multicolumn{4}{c}{Data-driven Method}\\ \cline{2-11}
 &\multicolumn{10}{c}{Train 6200  Test  100} \\ \cline{2-11}
 &NLCD \cite{qin2019infrared}  &PSTNN\cite{zhang2019infrared} &IPI\cite{gao2013infrared} &RIPT\cite{dai2017reweighted} &MGDWIE\cite{deng2016infrared} &NVMD\cite{wang2017infrared} & ACM\cite{dai2021asymmetric} & MDvsFA-cGAN\cite{wang2019miss} & TBC-Net\cite{zhao2019tbc} & UIU-Net \\
 \hline
IoU &0.1401&0.2295&0.0062&0.2019&0.0727&0.0533&  0.4208&0.3018 &0.4078&\bf 0.4773 \\
nIoU &0.2641&0.2875&0.2563&0.2523&0.2189&0.2460&  0.4127&0.2111  & 0.3710 & \bf 0.4721\\ 
\hline \hline
\multirow{2}{*}{Methods}&\multicolumn{10}{c}{Train 500  Test  100} \\ 
\cline{2-11}
&NLCD \cite{qin2019infrared}  &PSTNN\cite{zhang2019infrared} &IPI\cite{gao2013infrared} &RIPT\cite{dai2017reweighted} &MGDWIE\cite{deng2016infrared} &NVMD\cite{wang2017infrared} & ACM\cite{dai2021asymmetric} & MDvsFA-cGAN\cite{wang2019miss} & TBC-Net\cite{zhao2019tbc} & UIU-Net \\
\hline  
IoU &0.1401&0.2295&0.0062&0.2019&0.0727&0.0533&  0.2887&0.2596 &0.2320&\bf 0.4397 \\
nIoU &0.2641&0.2875&0.2563&0.2523&0.2189&0.2460&  0.3351&0.2088  & 0.3342 & \bf 0.4369\\ 
\bottomrule[1.5pt]
\end{tabular}
}
\label{tab:QuaSynthetic}
\end{table*}

\begin{figure*}[!t]
\centering\includegraphics[width=1.0\textwidth]{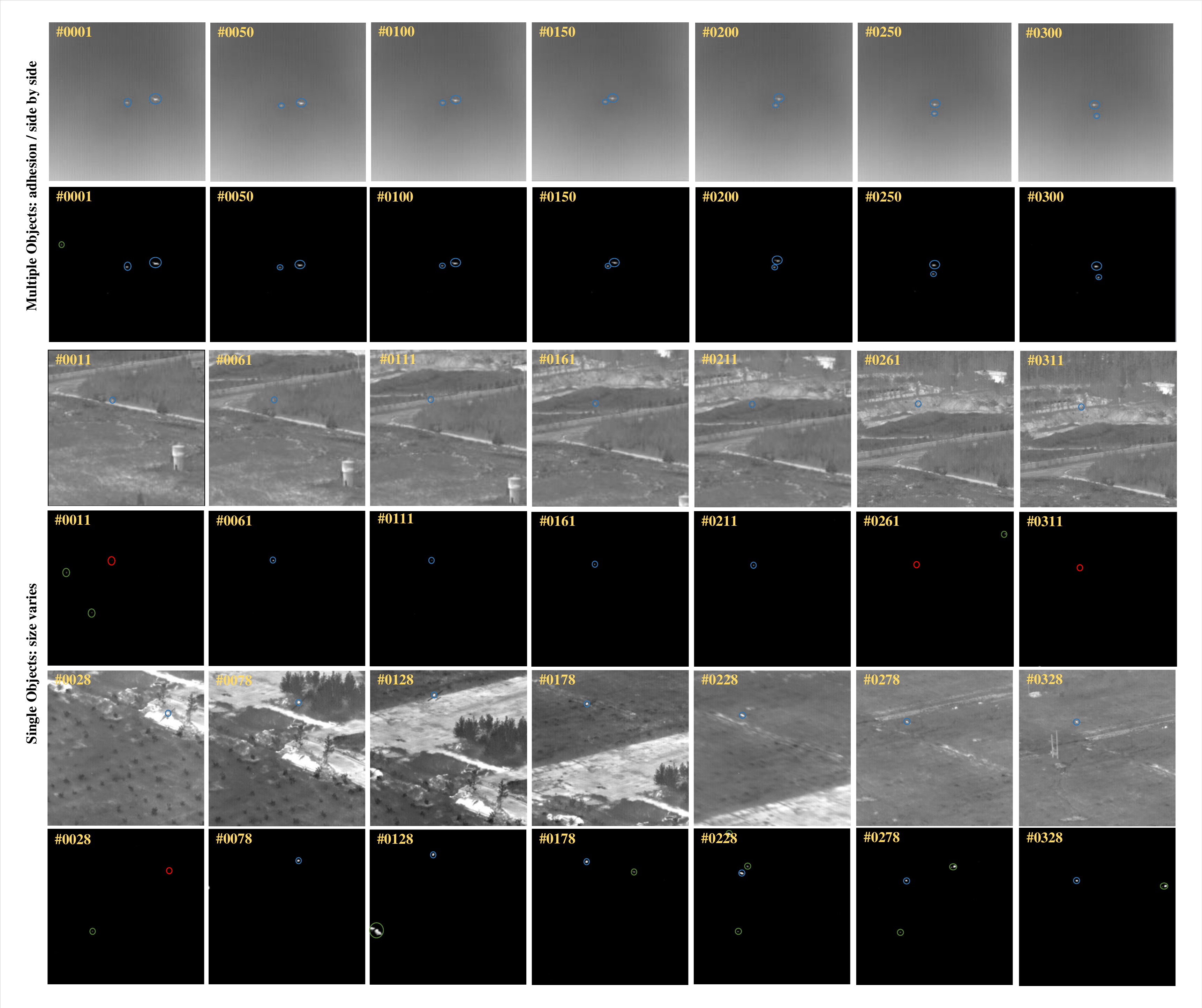}
\caption{Visual results of UIU-Net for ATR ground/air dataset. The blue, green, and red circles represent true, false, and missing detection, respectively. }
\label{fig:visGro}
\end{figure*}

\begin{figure}[!t]
    \centering
    \includegraphics[width=0.5\textwidth]{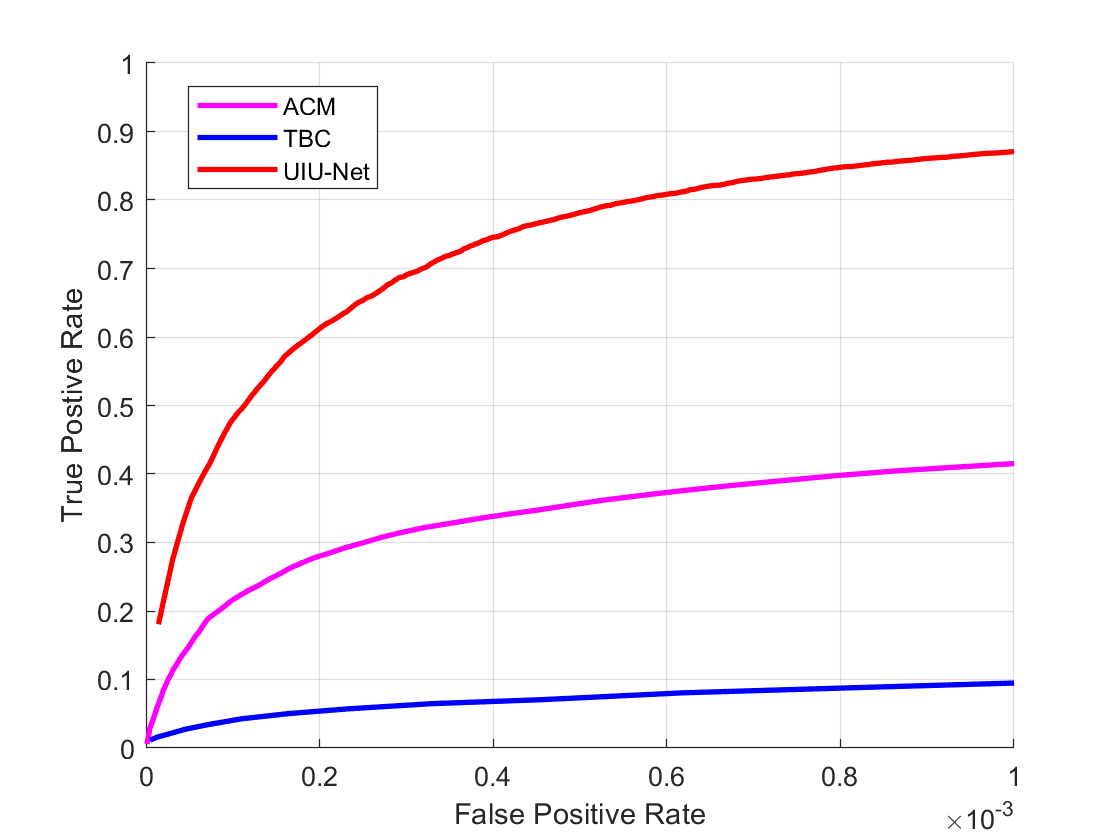}
    \caption{ROC Curve for the synthetic dataset. It's worth noting that all model-driven and MDvdFA\_cGAN ROC result is poor enough to be displayed.}
\label{fig:SynROC}
\end{figure}

\subsection{Results and Analysis on the Synthetic Data} \label{Sec:SynEx}
Similarly, the quantitative detection performance on the synthetic datasets are listed in Table \ref{tab:QuaSynthetic}, and the corresponding visualization results are shown in Fig. \ref{fig:visSys}. 

In addition, we have added an experiment to verify whether UIU-Net is overly dependent on the size of the dataset. To answer this question, two sub-experiments are designed. Specifically, 1) large-scale infrared training dataset: $6,200$ images in the configuration I am chosen and trained after manually removing the "poor" images. Here, the "poor" images are those that have null values or negative values after pre-processing. 2) small-scale infrared training dataset: $500$ images in 1) are randomly selected and reconstructed into a small sample data. similar to the SIRST data, the final detection output is the smallest outer rectangle of the optimal segmentation result, and the quantitative metrics for the comparing methods are IoU, nIoU, and ROC curves.


\textit{1) Quantitative Comparison.} To begin, we have given the longitudinal comparison between datasets. Compared to the real SIRST dataset as train data, the quantitative results for the synthetic dataset in Table \ref{tab:QuaSynthetic} has demonstrated the worse IoU and nIoU value. There are two main reasons for this: 1) resolution gap: the synthetic dataset has a resolution of $128\times 128$, which is lower than the SIRST dataset. Feature distinction and context representation ability are limited due to the lower image resolution. 2) reference labeling: the object labeling range in the synthetic data was erroneously inflated, and this labeling way is especially unfriendly to the quantitative evaluation of pixel-based methods. But the referenced labeling in the SIRST data is more closely matched to the real objects. 

Secondly, we have shown a data volume comparison. The ratio of large-scale to small-scale samples in the synthetic dataset is $12:1$. Although the IoU and nIoU for the proposed UIU-Nets have been decreased by about $4\%$, they are still optimal. Furthermore, the results from the both two small-scale datasets, including SIRST and small-scale synthetic data, show that UIU-Net is robust for small-scale data. It also indirectly indicates that UIU-Net is extremely friendly to practical applications.

Finally, the ROC curves for the above-comparison approaches are shown in Fig. \ref{fig:SynROC}, the proposed method still performs the best. Unfortunately, due to the complicated background and object variability, as well as insufficient image quality, none of the model-driven methods can be exhibited within the coordinate range. The detection performance of the MDvsFA-cGAN with underfitting for a small sample dataset is equally dismal. This emphasizes how important data volume and model fit are in data-driven methods.

\textit{2) Visual Comparison.} For the visualization results in Fig. \ref{fig:visSys}, the proposed UIU-Net has the highest similarity with the reference label for the object indicated in the blue rectangular box. However, as shown in the fourth line of Fig. \ref{fig:visSys}, some images have some missing detection even though the accuracy is assured. For other data-driven methods, the ACM output objects introduced erroneous bounds in comparison to its reference label. Due to the inaccuracy of the reference labels, MDvsFA-cGAN is unable to adequately balance the false alarm rate and missed alert rate, even with large-scale samples (as seen in the 
sixth column of Fig. \ref{fig:visSys}). For the model-driven methods, PSTNN achieves a rough object location, and MGDWIE has a higher missed detection, false detection, and lower accuracy.

\subsection{Generalization Analysis}
This section employs the detection model trained with SIRST true infrared small objects to verify the generalization capability of UIU-Net. It primarily consists of two parts of experiments.

\textit{1) Synthetic data:} As the same with section \ref{Sec:SynEx}, 100 test images have been selected to verify the generalization capability of UIU-Net, as shown in Table \ref{tab:gen}. It shows that the quantitative results of UIU-Net trained by $408$ SIRST data exceed the model trained by 500 synthetic data, although being lower than the results of training by $6,200$ synthetic data. In addition, it exceeds the other compared methods in Table \ref{tab:QuaSynthetic}, no matter $6,200$ or $500$ trained data. 

\begin{table}[!t]
\renewcommand\arraystretch{1.5}
\centering
\tiny
\caption{Quantitative comparison of UIU-Net under different train data. The best results are shown in bold.}
\resizebox{0.5\textwidth}{!}{
\begin{tabular}{c|c|c|c|c}
\toprule[1.0pt]
Model &Train/Num & Test/Num &IoU &nIoU \\
\hline
UIU-Net & SIRST/407& Synthetic/100 & 0.4521 & \bf 0.4740 \\
UIU-Net & Synthetic/6,200& Synthetic/100 & \bf 0.4773 &  0.4721 \\
UIU-Net & Synthetic/500 & Synthetic/100 & 0.4397 & 0.4369\\
\bottomrule[1.0pt]
\end{tabular}
}
\label{tab:gen}
\end{table}

\textit{2) ATR Sequential Data:} Due to the shortages of public single-frame infrared small object datasets, we employ ATR sequential data as the test data to verify the generalization capability of UIU-Net. This inspired a question: how does the proposed UIU-Net perform in terms of the video sequence dataset? We sample data2, data6, and data7 segments for visualization validation due to the limits of the center point label, as shown in Fig. \ref{fig:visGro}, where we display a detection result every 50 frames. For data2, UIU-Net has good detection performance. All visualization results except for $\#0001$ are well for cross-flying UAVs in a close and sky background (see rows 1-2 in Fig. \ref{fig:visGro}). Due to the intricacy of the ground background and the failure to consider timeliness, the missed detection is more problematic for the UAV flying near the original ground backdrop in data6 (see the rows 3-4 in Fig. \ref{fig:visGro}). In contrast to the previous two data segments, the ground change of data7 will invariably contain flash elements, interfering with the UAV detection (see rows 5-6 in Fig. \ref{fig:visGro}). 

To sum up, the UIU-Net model trained on the SIRST dataset shows superior generalization performance for small object detection in three video data segments. Currently, there are some other works in the field of saliency object detection, e.g., Contrast based filtering \cite{perazzi2012saliency}, Context-aware saliency \cite{5539929}, spectral residual method \cite{hou2007saliency}, etc., which can also achieve the high generalization ability in the video object detection.

Currently, detecting unknown scenes or objects is still a challenging and noteworthy research problem. There have been some works, e.g., background subtraction\cite{tezcan2020bsuv}, scene independent end-to-end spatiotemporal feature learning framework\cite{mandal20203dcd}, meta-knowledge learning and domain adaptation\cite{zhang2021meta}, achieving the high robustness and generalization ability for unseen scenarios. Actually, the experiments shown in Fig. \ref{fig:visGro} are a case of object detection in the unseen video dataset (ATR ground/air video dataset). More specifically, the ATR dataset is an infrared detection and tracking dataset with one or more fixed-wing UAV objects that differ in size, interference, storage and mobility, and other properties (\textit{cf.} the SIRST dataset). This naturally brings more unseen objects to be detected. Moreover, the image's background is diverse, including the sky, the earth, and so on, leading to more unseen scenes as well (\textit{cf.} the SIRST dataset).

\section{Conclusion}
In this paper, we propose an interactive-cross attention-nested U-Net network, called UIU-Net. The UIU-Net increases the network depth without reducing object resolution, and it does not rely on the classification backbone to avoid information loss of small objects during downsampling. Furthermore, by including an interactive-across attention module, deep multi-scale features can be encoded while object global and local context representation is improved. Quantitative experiments on SIRST and synthetic datasets, as well as generalization studies on ATR ground/sir datasets, demonstrate the superiority and efficiency of UIU-Net. UIU-Net brings a new perspective on infrared small object detection, and its robustness and stability make it a popular choice for real-world applications. As a result, future work will concentrate on fine-tuning the network to increase object detection accuracy and efficiency, especially in complex video sequences. 

As sensors improved, the feasibility and ubiquity of multi-source data acquisition have expanded. Objects in multi-source data, such as visible and infrared data, not only have various attributes but also have information that intersects and complements one another. Small object detection based on deep learning methods for multi-source data fusion \cite{fusion1,fusion2} will be investigated in future work to improve the performance of object detection. In addition, we will also focus on the key factors that influence the model's capability to adjust to various scenarios, and further improve the model's generalization ability to unknown scenarios.


\section*{Acknowledgment}
The authors would like to thank the Nanjing University of Aeronautics and Astronautics for providing the SIRST dataset, the Nanjing University of Science \& Technology for providing the Synthetic dataset, and the National University of Defense Technology for providing the ground/air background data.

\bibliographystyle{ieeetr}
\bibliography{reference}

\begin{thebibliography}{10}

\bibitem{Sta2016}
T.~R. Goodall, A.~C. Bovik, and N.~G. Paulter, ``Tasking on natural statistics
  of infrared images,'' {\em IEEE Transactions on Image Processing}, vol.~25,
  no.~1, pp.~65--79, 2016.

\bibitem{wu2022deep}
X.~Wu, W.~Li, D.~Hong, R.~Tao, and Q.~Du, ``Deep learning for unmanned aerial
  vehicle-based object detection and tracking: a survey,'' {\em IEEE Geoscience
  and Remote Sensing Magazine}, vol.~10, no.~1, pp.~91--124, 2022.

\bibitem{LSK2017}
S.~K. Biswas and P.~Milanfar, ``Linear support tensor machine with lsk
  channels: Pedestrian detection in thermal infrared images,'' {\em IEEE
  Transactions on Image Processing}, vol.~26, no.~9, pp.~4229--4242, 2017.

\bibitem{CGAN2019}
P.~Wang and X.~Bai, ``Thermal infrared pedestrian segmentation based on
  conditional gan,'' {\em IEEE Transactions on Image Processing}, vol.~28,
  no.~12, pp.~6007--6021, 2019.

\bibitem{TNLRS2020}
H.~Zhu, H.~Ni, S.~Liu, G.~Xu, and L.~Deng, ``Tnlrs: Target-aware non-local
  low-rank modeling with saliency filtering regularization for infrared small
  target detection,'' {\em IEEE Transactions on Image Processing}, vol.~29,
  pp.~9546--9558, 2020.

\bibitem{bae2012edge}
T.-W. Bae, F.~Zhang, and I.-S. Kweon, ``Edge directional 2d lms filter for
  infrared small target detection,'' {\em Infrared Physics \& Technology},
  vol.~55, no.~1, pp.~137--145, 2012.

\bibitem{gao2013infrared}
C.~Gao, D.~Meng, Y.~Yang, Y.~Wang, X.~Zhou, and A.~G. Hauptmann, ``Infrared
  patch-image model for small target detection in a single image,'' {\em IEEE
  transactions on image processing}, vol.~22, no.~12, pp.~4996--5009, 2013.

\bibitem{chen2013local}
C.~P. Chen, H.~Li, Y.~Wei, T.~Xia, and Y.~Y. Tang, ``A local contrast method
  for small infrared target detection,'' {\em IEEE transactions on geoscience
  and remote sensing}, vol.~52, no.~1, pp.~574--581, 2013.

\bibitem{wei2016multiscale}
Y.~Wei, X.~You, and H.~Li, ``Multiscale patch-based contrast measure for small
  infrared target detection,'' {\em Pattern Recognition}, vol.~58,
  pp.~216--226, 2016.

\bibitem{he2015small}
Y.~He, M.~Li, J.~Zhang, and Q.~An, ``Small infrared target detection based on
  low-rank and sparse representation,'' {\em Infrared Physics \& Technology},
  vol.~68, pp.~98--109, 2015.

\bibitem{deng2016infrared}
H.~Deng, X.~Sun, M.~Liu, C.~Ye, and X.~Zhou, ``Infrared small-target detection
  using multiscale gray difference weighted image entropy,'' {\em IEEE
  Transactions on Aerospace and Electronic Systems}, vol.~52, no.~1,
  pp.~60--72, 2016.

\bibitem{qin2019infrared}
Y.~Qin, L.~Bruzzone, C.~Gao, and B.~Li, ``Infrared small target detection based
  on facet kernel and random walker,'' {\em IEEE Transactions on Geoscience and
  Remote Sensing}, vol.~57, no.~9, pp.~7104--7118, 2019.

\bibitem{zhang2019infrared}
L.~Zhang and Z.~Peng, ``Infrared small target detection based on partial sum of
  the tensor nuclear norm,'' {\em Remote Sensing}, vol.~11, no.~4, p.~382,
  2019.

\bibitem{dai2017reweighted}
Y.~Dai and Y.~Wu, ``Reweighted infrared patch-tensor model with both nonlocal
  and local priors for single-frame small target detection,'' {\em IEEE journal
  of selected topics in applied earth observations and remote sensing},
  vol.~10, no.~8, pp.~3752--3767, 2017.

\bibitem{wang2017infrared}
X.~Wang, Z.~Peng, P.~Zhang, and Y.~He, ``Infrared small target detection via
  nonnegativity-constrained variational mode decomposition,'' {\em IEEE
  Geoscience and Remote Sensing Letters}, vol.~14, no.~10, pp.~1700--1704,
  2017.

\bibitem{hong2021graph}
D.~Hong, L.~Gao, J.~Yao, B.~Zhang, P.~Antonio, and J.~Chanussot, ``Graph
  convolutional networks for hyperspectral image classification,'' {\em IEEE
  Transactions on Geoscience and Remote Sensing}, vol.~59, pp.~5966--5978, Jul.
  2021.

\bibitem{hong2021more}
D.~Hong, L.~Gao, N.~Yokoya, J.~Yao, J.~Chanussot, Q.~Du, and B.~Zhang, ``More
  diverse means better: Multimodal deep learning meets remote-sensing imagery
  classification,'' {\em IEEE Transactions on Geoscience and Remote Sensing},
  vol.~59, pp.~4340--4354, May 2021.

\bibitem{wang2017small}
W.~Wang, H.~Qin, W.~Cheng, C.~Wang, H.~Leng, and H.~Zhou, ``Small target
  detection in infrared image using convolutional neural networks,'' in {\em
  AOPC 2017: Optical Sensing and Imaging Technology and Applications},
  vol.~10462, p.~1046250, International Society for Optics and Photonics, 2017.

\bibitem{nasrabadi2019deeptarget}
N.~M. Nasrabadi, ``Deeptarget: An automatic target recognition using deep
  convolutional neural networks,'' {\em IEEE Transactions on Aerospace and
  Electronic Systems}, vol.~55, no.~6, pp.~2687--2697, 2019.

\bibitem{fan2018dim}
Z.~Fan, D.~Bi, L.~Xiong, S.~Ma, L.~He, and W.~Ding, ``Dim infrared image
  enhancement based on convolutional neural network,'' {\em Neurocomputing},
  vol.~272, pp.~396--404, 2018.

\bibitem{liangkui2018using}
L.~Liangkui, W.~Shaoyou, and T.~Zhongxing, ``Using deep learning to detect
  small targets in infrared oversampling images,'' {\em Journal of Systems
  Engineering and Electronics}, vol.~29, no.~5, pp.~947--952, 2018.

\bibitem{wang2019detection}
K.~Wang, S.~Li, S.~Niu, and K.~Zhang, ``Detection of infrared small targets
  using feature fusion convolutional network,'' {\em IEEE Access}, vol.~7,
  pp.~146081--146092, 2019.

\bibitem{mcintosh2020infrared}
B.~McIntosh, S.~Venkataramanan, and A.~Mahalanobis, ``Infrared target detection
  in cluttered environments by maximization of a target to clutter ratio (tcr)
  metric using a convolutional neural network,'' {\em IEEE Transactions on
  Aerospace and Electronic Systems}, vol.~57, no.~1, pp.~485--496, 2020.

\bibitem{hou2021ristdnet}
Q.~Hou, Z.~Wang, F.~Tan, Y.~Zhao, H.~Zheng, and W.~Zhang, ``Ristdnet: Robust
  infrared small target detection network,'' {\em IEEE Geoscience and Remote
  Sensing Letters}, 2021.

\bibitem{zhao2020novel}
B.~Zhao, C.~Wang, Q.~Fu, and Z.~Han, ``A novel pattern for infrared small
  target detection with generative adversarial network,'' {\em IEEE
  Transactions on Geoscience and Remote Sensing}, vol.~59, no.~5,
  pp.~4481--4492, 2020.

\bibitem{hsieh2021fast}
T.-H. Hsieh, C.-L. Chou, Y.-P. Lan, P.-H. Ting, and C.-T. Lin, ``Fast and
  robust infrared image small target detection based on the convolution of
  layered gradient kernel,'' {\em IEEE Access}, vol.~9, pp.~94889--94900, 2021.

\bibitem{zhao2019tbc}
M.~Zhao, L.~Cheng, X.~Yang, P.~Feng, L.~Liu, and N.~Wu, ``Tbc-net: A real-time
  detector for infrared small target detection using semantic constraint,''
  {\em arXiv preprint arXiv:2001.05852}, 2019.

\bibitem{strudel2021segmenter}
R.~Strudel, R.~Garcia, I.~Laptev, and C.~Schmid, ``Segmenter: Transformer for
  semantic segmentation,'' in {\em Proceedings of the IEEE/CVF International
  Conference on Computer Vision}, pp.~7262--7272, 2021.

\bibitem{zhang2021automatic}
D.~Zhang, J.~Zhang, Q.~Zhang, J.~Han, S.~Zhang, and J.~Han, ``Automatic
  pancreas segmentation based on lightweight dcnn modules and spatial prior
  propagation,'' {\em Pattern Recognition}, vol.~114, p.~107762, 2021.

\bibitem{dai2021asymmetric}
Y.~Dai, Y.~Wu, F.~Zhou, and K.~Barnard, ``Asymmetric contextual modulation for
  infrared small target detection,'' in {\em Proceedings of the IEEE/CVF Winter
  Conference on Applications of Computer Vision}, pp.~950--959, 2021.

\bibitem{dai2021attentional}
Y.~Dai, Y.~Wu, F.~Zhou, and K.~Barnard, ``Attentional local contrast networks
  for infrared small target detection,'' {\em IEEE Transactions on Geoscience
  and Remote Sensing}, 2021.

\bibitem{li2021dense}
B.~Li, C.~Xiao, L.~Wang, Y.~Wang, Z.~Lin, M.~Li, W.~An, and Y.~Guo, ``Dense
  nested attention network for infrared small target detection,'' {\em arXiv
  preprint arXiv:2106.00487}, 2021.

\bibitem{hong2021multimodal}
D.~Hong, J.~Yao, D.~Meng, Z.~Xu, and J.~Chanussot, ``Multimodal gans: Toward
  crossmodal hyperspectral--multispectral image segmentation,'' {\em IEEE
  Transactions on Geoscience and Remote Sensing}, vol.~59, no.~6,
  pp.~5103--5113, 2021.

\bibitem{long2015fully}
J.~Long, E.~Shelhamer, and T.~Darrell, ``Fully convolutional networks for
  semantic segmentation,'' in {\em Proceedings of the IEEE conference on
  computer vision and pattern recognition}, pp.~3431--3440, 2015.

\bibitem{badrinarayanan2017segnet}
V.~Badrinarayanan, A.~Kendall, and R.~Cipolla, ``Segnet: A deep convolutional
  encoder-decoder architecture for image segmentation,'' {\em IEEE transactions
  on pattern analysis and machine intelligence}, vol.~39, no.~12,
  pp.~2481--2495, 2017.

\bibitem{wu2019fastfcn}
H.~Wu, J.~Zhang, K.~Huang, K.~Liang, and Y.~Yu, ``Fastfcn: Rethinking dilated
  convolution in the backbone for semantic segmentation,'' {\em arXiv preprint
  arXiv:1903.11816}, 2019.

\bibitem{li2019scale}
Y.~Li, Y.~Chen, N.~Wang, and Z.~Zhang, ``Scale-aware trident networks for
  object detection,'' in {\em Proceedings of the IEEE/CVF International
  Conference on Computer Vision}, pp.~6054--6063, 2019.

\bibitem{singh2018analysis}
B.~Singh and L.~S. Davis, ``An analysis of scale invariance in object detection
  snip,'' in {\em Proceedings of the IEEE conference on computer vision and
  pattern recognition}, pp.~3578--3587, 2018.

\bibitem{vaswani2017attention}
A.~Vaswani, N.~Shazeer, N.~Parmar, J.~Uszkoreit, L.~Jones, A.~N. Gomez,
  {\L}.~Kaiser, and I.~Polosukhin, ``Attention is all you need,'' in {\em
  Advances in neural information processing systems}, pp.~5998--6008, 2017.

\bibitem{guo2021beyond}
M.-H. Guo, Z.-N. Liu, T.-J. Mu, and S.-M. Hu, ``Beyond self-attention: External
  attention using two linear layers for visual tasks,'' {\em arXiv preprint
  arXiv:2105.02358}, 2021.

\bibitem{hu2018squeeze}
J.~Hu, L.~Shen, and G.~Sun, ``Squeeze-and-excitation networks,'' in {\em
  Proceedings of the IEEE conference on computer vision and pattern
  recognition}, pp.~7132--7141, 2018.

\bibitem{woo2018cbam}
S.~Woo, J.~Park, J.-Y. Lee, and I.~S. Kweon, ``Cbam: Convolutional block
  attention module,'' in {\em Proceedings of the European conference on
  computer vision (ECCV)}, pp.~3--19, 2018.

\bibitem{park2018bam}
J.~Park, S.~Woo, J.-Y. Lee, and I.~S. Kweon, ``Bam: Bottleneck attention
  module,'' {\em arXiv preprint arXiv:1807.06514}, 2018.

\bibitem{qin2020u2}
X.~Qin, Z.~Zhang, C.~Huang, M.~Dehghan, O.~R. Zaiane, and M.~Jagersand,
  ``U2-net: Going deeper with nested u-structure for salient object
  detection,'' {\em Pattern Recognition}, vol.~106, p.~107404, 2020.

\bibitem{wang2019miss}
H.~Wang, L.~Zhou, and L.~Wang, ``Miss detection vs. false alarm: Adversarial
  learning for small object segmentation in infrared images,'' in {\em
  Proceedings of the IEEE/CVF International Conference on Computer Vision},
  pp.~8509--8518, 2019.

\bibitem{hui2020data}
H.~Bingwei, S.~Zhiyong, F.~Hongqi, Z.~Ping, H.~Weidong, Z.~Xiaofeng,
  L.~Jianguo, S.~Hongyan, J.~Wei, Z.~Yongjie, and B.~Yaxi, ``A dataset for
  infrared detection and tracking of dim-small aircraft targets under ground /
  air background,'' {\em China Scientific Data: online version in English and
  Chinese}, vol.~5, no.~3, p.~12, 2020.

\bibitem{perazzi2012saliency}
F.~Perazzi, P.~Kr{\"a}henb{\"u}hl, Y.~Pritch, and A.~Hornung, ``Saliency
  filters: Contrast based filtering for salient region detection,'' in {\em
  2012 IEEE conference on computer vision and pattern recognition},
  pp.~733--740, IEEE, 2012.

\bibitem{5539929}
S.~Goferman, L.~Zelnik-Manor, and A.~Tal, ``Context-aware saliency detection,''
  in {\em 2010 IEEE Computer Society Conference on Computer Vision and Pattern
  Recognition}, pp.~2376--2383, 2010.

\bibitem{hou2007saliency}
X.~Hou and L.~Zhang, ``Saliency detection: A spectral residual approach,'' in
  {\em 2007 IEEE Conference on computer vision and pattern recognition},
  pp.~1--8, Ieee, 2007.

\bibitem{tezcan2020bsuv}
O.~Tezcan, P.~Ishwar, and J.~Konrad, ``Bsuv-net: A fully-convolutional neural
  network for background subtraction of unseen videos,'' in {\em Proceedings of
  the IEEE/CVF Winter Conference on Applications of Computer Vision},
  pp.~2774--2783, 2020.

\bibitem{mandal20203dcd}
M.~Mandal, V.~Dhar, A.~Mishra, S.~K. Vipparthi, and M.~Abdel-Mottaleb, ``3dcd:
  Scene independent end-to-end spatiotemporal feature learning framework for
  change detection in unseen videos,'' {\em IEEE Transactions on Image
  Processing}, vol.~30, pp.~546--558, 2020.

\bibitem{zhang2021meta}
J.~Zhang, X.~Zhang, Y.~Zhang, Y.~Duan, Y.~Li, and Z.~Pan, ``Meta-knowledge
  learning and domain adaptation for unseen background subtraction,'' {\em IEEE
  Transactions on Image Processing}, vol.~30, pp.~9058--9068, 2021.

\bibitem{fusion1}
H.~Li, X.-J. Wu, and J.~Kittler, ``Mdlatlrr: A novel decomposition method for
  infrared and visible image fusion,'' {\em IEEE Transactions on Image
  Processing}, vol.~29, pp.~4733--4746, 2020.

\bibitem{fusion2}
W.~Liang, G.~Wang, J.~Lai, and X.~Xie, ``Homogeneous-to-heterogeneous:
  Unsupervised learning for rgb-infrared person re-identification,'' {\em IEEE
  Transactions on Image Processing}, vol.~30, pp.~6392--6407, 2021.

\end{thebibliography}

\begin{IEEEbiography}[{\includegraphics[width=1in,height=1.25in,clip,keepaspectratio]{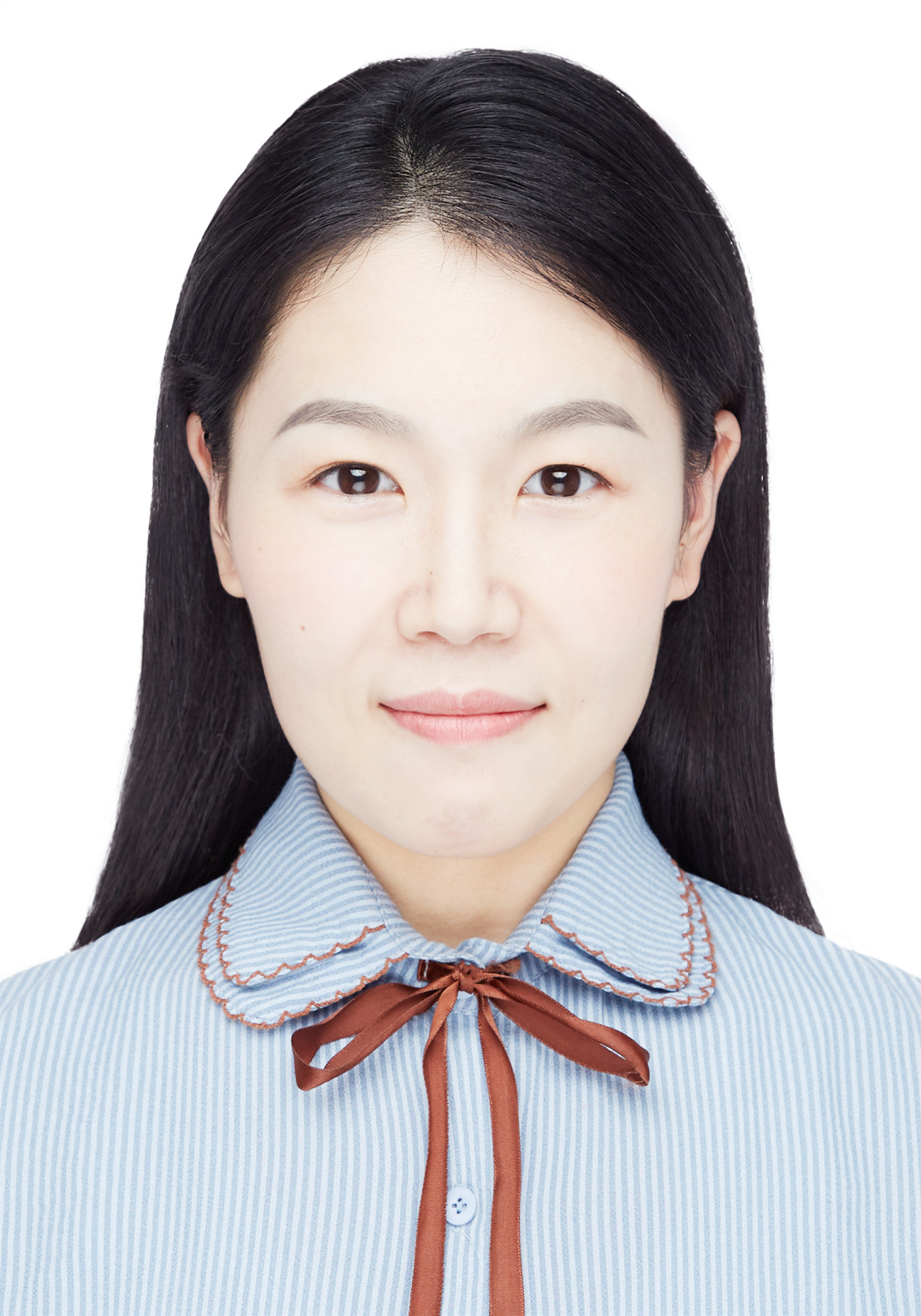}}]{Xin Wu} (IEEE Member) received the M.Sc. degree in Computer Science and Technology from the College of Information Engineering, Qingdao University, Qingdao, China, in 2014, the Ph.D. degree from the School of Information and Electronics, Beijing Institute of Technology (BIT), Beijing, China, in 2020. She is currently an Assistant Professor in the School of Computer Science, Beijing University of Posts and Telecommunications (BUPT), Beijing, China. Her research interests include signal/image processing, fractional Fourier transform, deep learning, and their applications in biometrics and geospatial object detection.

She was a recipient of the Best Reviewer Award of the IEEE JSTARS in 2022 and the Jose Bioucas Dias award for recognizing the outstanding paper at WHISPERS in 2021.
\end{IEEEbiography}

\vskip -2\baselineskip plus -1fil

\begin{IEEEbiography}[{\includegraphics[width=1in,height=1.25in,clip,keepaspectratio]{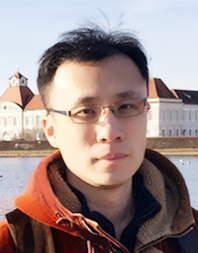}}]{Danfeng Hong} (IEEE Senior Member) received the M.Sc. degree (summa cum laude) in computer vision from the College of Information Engineering, Qingdao University, Qingdao, China, in 2015, the Dr. -Ing degree (summa cum laude) from the Signal Processing in Earth Observation (SiPEO), Technical University of Munich (TUM), Munich, Germany, in 2019. 

He is currently a Professor with the Key Laboratory of Computational Optical Imaging Technology, Aerospace Information Research Institute, Chinese Academy of Sciences (CAS). Before joining CAS, he has been a Research Scientist and led a Spectral Vision Working Group at the Remote Sensing Technology Institute (IMF), German Aerospace Center (DLR), Oberpfaffenhofen, Germany. He was also an Adjunct Scientist at GIPSA-lab, Grenoble INP, CNRS, Univ. Grenoble Alpes, Grenoble, France. His research interests include signal/image processing, hyperspectral remote sensing, machine / deep learning, artificial intelligence, and their applications in Earth Vision.

Dr. Hong is an Associate Editor for the IEEE Transactions on Geoscience and Remote Sensing (TGRS), an Editorial Board Member of Remote Sensing, an Editorial Advisory Board Member of ISPRS Journal of Photogrammetry and Remote Sensing. He was a recipient of the Best Reviewer Award of the IEEE TGRS in 2021 and 2022, and the Best Reviewer Award of the IEEE JSTARS in 2022, the Jose Bioucas Dias Award for recognizing the outstanding paper at WHISPERS in 2021, the Remote Sensing Young Investigator Award in 2022, the IEEE GRSS Early Career Award in 2022, and a Highly Cited Researcher (Clarivate Analytics/Thomson Reuters) in 2022.
\end{IEEEbiography}

\vskip -2\baselineskip plus -1fil

\begin{IEEEbiography}[{\includegraphics[width=1in,height=1.25in,clip,keepaspectratio]{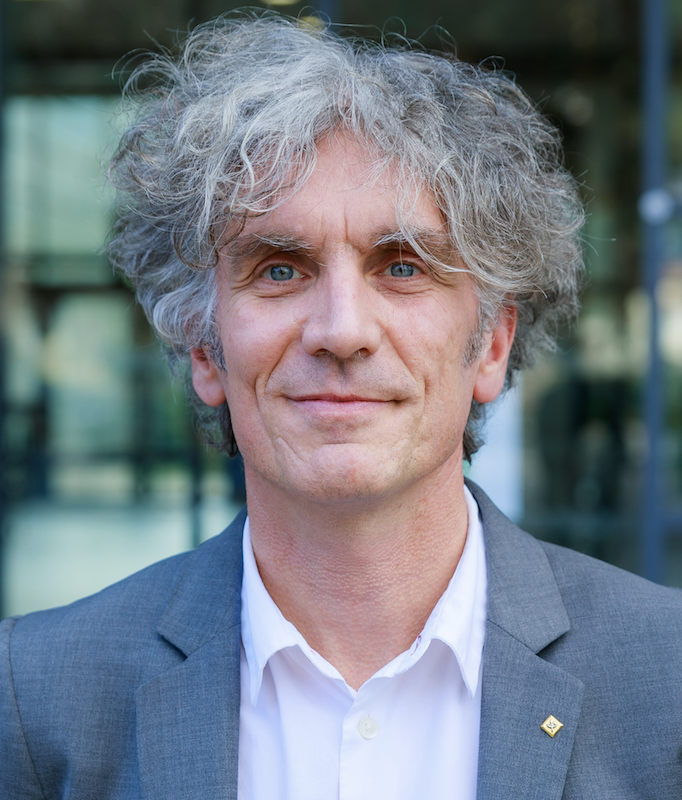}}]{Jocelyn Chanussot}
(M'04--SM'04--F'12) received the M.Sc. degree in electrical engineering from the Grenoble Institute of Technology (Grenoble INP), Grenoble, France, in 1995, and the Ph.D. degree from the Université de Savoie, Annecy, France, in 1998. Since 1999, he has been with Grenoble INP, where he is currently a Professor of signal and image processing. His research interests include image analysis, hyperspectral remote sensing, data fusion, machine learning and artificial intelligence. He has been a visiting scholar at Stanford University (USA), KTH (Sweden) and NUS (Singapore). Since 2013, he is an Adjunct Professor of the University of Iceland. In 2015-2017, he was a visiting professor at the University of California, Los Angeles (UCLA). He holds the AXA chair in remote sensing and is an Adjunct professor at the Chinese Academy of Sciences, Aerospace Information research Institute, Beijing.

Dr. Chanussot is the founding President of IEEE Geoscience and Remote Sensing French chapter (2007-2010) which received the 2010 IEEE GRS-S Chapter Excellence Award. He has received multiple outstanding paper awards. He was the Vice-President of the IEEE Geoscience and Remote Sensing Society, in charge of meetings and symposia (2017-2019). He was the General Chair of the first IEEE GRSS Workshop on Hyperspectral Image and Signal Processing, Evolution in Remote sensing (WHISPERS). He was the Chair (2009-2011) and  Cochair of the GRS Data Fusion Technical Committee (2005-2008). He was a member of the Machine Learning for Signal Processing Technical Committee of the IEEE Signal Processing Society (2006-2008) and the Program Chair of the IEEE International Workshop on Machine Learning for Signal Processing (2009). He is an Associate Editor for the IEEE Transactions on Geoscience and Remote Sensing and the Proceedings of the IEEE. He was the Editor-in-Chief of the IEEE Journal of Selected Topics in Applied Earth Observations and Remote Sensing (2011-2015) and an Associate Editor for IEEE Transactions on Image Processing. In 2014 he served as a Guest Editor for the IEEE Signal Processing Magazine. He is a Fellow of the IEEE, a member of the Institut Universitaire de France (2012-2017) and a Highly Cited Researcher (Clarivate Analytics/Thomson Reuters).
\end{IEEEbiography}

\end{document}